\pdfoutput=1
\documentclass[10pt,reqno]{amsart}
\usepackage{graphicx}
\usepackage{amsaddr}
\usepackage{multirow}
\usepackage{siunitx}
\usepackage{tikz}
\usepackage{booktabs}

\usetikzlibrary{positioning, arrows.meta, fit, backgrounds, calc}

\usepackage{amsmath}
\usepackage{amssymb}
\usepackage{amsthm}
\usepackage{amscd}
\usepackage{amsfonts}
\usepackage{mathtools}
\usepackage{xcolor}
\usepackage{cancel}
\usepackage{graphicx}
\usepackage{hyperref}
\usepackage[normalem]{ulem}
\usepackage{float}
\usepackage{cancel}
\usepackage[
backend=biber,
style=alphabetic,
sorting=ynt,
maxbibnames=999
]{biblatex}
\renewbibmacro{in:}{%
  \ifentrytype{article}
    {}
    {\printtext{\bibstring{in}\intitlepunct}}}
\usepackage{subcaption}
\usepackage[linesnumbered,boxruled,lined,noend,ruled,vlined]{algorithm2e}
\SetAlFnt{\footnotesize}
\SetKw{Break}{break}
\SetKwComment{tcp}{$\triangleright$\ }{}
\usepackage{xcolor,paralist,hyperref,fancyhdr,etoolbox}
\usepackage[indentafter]{titlesec}

\usepackage{setspace}
\usepackage{todonotes}

\usepackage{enumitem}

\usepackage[export]{adjustbox}

\usepackage{indentfirst,csquotes}

\usepackage{tikz}
\usetikzlibrary{shapes, arrows, positioning, fit, decorations.pathreplacing}

\newtheorem{theorem}{Theorem}[]
\newtheorem{assumption}{Assumption}

\newtheorem{lemma}[theorem]{Lemma}

\newtheorem{corollary}[theorem]{Corollary}

\newtheorem{remark}[theorem]{Remark}

\newcommand{\norm}[1]{\left\| #1 \right\|}

\titleformat{name=\section}{}{\thetitle.}{0.8em}{\centering\scshape}
\titleformat{name=\subsection}[runin]{}{\thetitle.}{0.5em}{\bfseries}[.]
\titleformat{name=\subsubsection}[runin]{}{\thetitle.}{0.5em}{\itshape}[.]
\titleformat{name=\paragraph,numberless}[runin]{}{}{0em}{}[.]
\titlespacing{\paragraph}{0em}{0em}{0.5em}
\titleformat{name=\subparagraph,numberless}[runin]{}{}{0em}{}[.]
\titlespacing{\subparagraph}{0em}{0em}{0.5em}

\hypersetup{ colorlinks=true, linkcolor=black, filecolor=black, urlcolor=black }

\usepackage{lipsum}

\DeclareMathOperator*{\argmin}{arg min}

\newcommand{\R}[0]{\mathbb{R}}
\newcommand{\E}[0]{\mathbb{E}}
\renewcommand{\L}{{\mathbb{L}}}
\renewcommand{\P}[0]{\mathbb{P}}

\newcommand{\B}[0]{\mathcal{B}}

\newcommand{\F}[0]{\mathcal{F}}
\newcommand{\G}[0]{\mathcal{G}}

\newcommand{\mfe}{{\mathfrak{e}}}
\newcommand{\mfn}{{\mathfrak{n}}}
\newcommand{\mfY}{{\mathfrak{Y}}}
\newcommand{\mfS}{{\mathfrak{S}}}
\newcommand{\mfs}{{\mathfrak{s}}}

\newcommand{\ds}{{{\rm d} s}}
\newcommand{\du}{{{\rm d} u}}
\newcommand{\dt}{{{\rm d} t}}
\newcommand{\dW}{{{\rm d} W}}
\newcommand{\dX}{{{\rm d} X}}
\newcommand{\dB}{{{\rm d} B}}
\newcommand{\dY}{{{\rm d} Y}}
\newcommand{\dK}{{{\rm d} K}}

\newcommand{\T}{{\mathcal{T}}}
\newcommand{\tT}{{t\in\mathcal{T}}}

\newcommand{\enet}{\ensuremath{\varepsilon_{\scriptscriptstyle\mathrm{NN}}}}

\begin{document}

\title{Deep Learning and Elicitability for McKean-Vlasov FBSDEs With Common Noise} 
\author[Felipe Antunes]{Felipe J. P. Antunes}
\address{School of Applied Mathematics, Getulio Vargas Foundation, Brazil}
\email{felipe.antunes@fgv.br}

\author[Yuri F. Saporito]{Yuri F. Saporito}
\address{School of Applied Mathematics, Getulio Vargas Foundation, Brazil}
\email{yuri.saporito@fgv.br}

\author[Sebastian Jaimungal]{Sebastian Jaimungal}
\address{Department of Statistical Sciences, University of Toronto, Canada and \\
Oxford-Man Institute for Quantitative Finance, University of Oxford}
\email{sebastian.jaimungal@utoronto.ca}

\maketitle

\begin{abstract}
We present a novel numerical method for solving McKean-Vlasov forward-backward stochastic differential equations (MV-FBSDEs) with common noise, combining Picard iterations, elicitability and deep learning. The key innovation involves elicitability to derive a path-wise loss function, enabling efficient training of neural networks to approximate both the backward process and the conditional expectations arising from common noise -- without requiring computationally expensive nested Monte Carlo simulations. The mean-field interaction term is parameterized via a recurrent neural network trained to minimize an elicitable score, while the backward process is approximated through a hybrid feedforward and recurrent network representing the decoupling field. We validate the algorithm on a systemic risk inter-bank borrowing and lending model, where analytical solutions exist, demonstrating accurate recovery of the true solution. We further extend the model to quantile--mediated interactions, showcasing the flexibility of the elicitability framework beyond conditional means or moments. Finally, we apply the method to a non-stationary Aiyagari–Bewley–Huggett economic growth model with endogenous interest rates, illustrating its applicability to complex mean-field games without closed-form solutions.
\end{abstract}

\section{Introduction}

McKean--Vlasov forward--backward stochastic differential equations
(MV-FBSDEs) arise naturally in stochastic control problems where the
dynamics or cost functionals depend on the law of the solution process.
They combine the forward--backward structure characteristic of the
stochastic maximum principle with mean-field interactions through the
law of the process: a forward diffusion $(X_t)_{t\in[0,T]}$ is coupled
with backward processes $(Y_t, Z_t)_{t\in[0,T]}$, and the coefficients
depend on both the current state and the law of $X_t$ -- typically,
though not always, through its conditional moments. The coupling
between the forward and backward components, together with the
dependence on the law of the solution, makes these equations
challenging from both theoretical and numerical perspectives. Of
particular interest is the case in which the system is subject to
common noise, where the mean-field interaction acquires additional
structure through conditioning on a shared source of randomness. Such
systems play a fundamental role in the probabilistic analysis of
mean-field games and mean-field-type control problems, where they
characterize Nash equilibria and optimal controls, respectively
\cite{carmona2018probabilistic}.

This paper proposes a numerical method that addresses the two principal
difficulties of MV-FBSDEs with common noise: the forward--backward
coupling and the dependence on the stochastic measure flow. The
coupling is addressed by an outer Picard iteration over the entire time
horizon, alternating between a forward step (solving the SDE with the
backward components frozen) and a backward step (solving the BSDE with
the forward components frozen). The measure flow is handled through
\emph{elicitability} \cite{fissler2016higher}:
 any statistic of the conditional law
that minimizes a suitable scoring function -- conditional means,
moments, quantiles, or jointly elicitable pairs such as quantile and
expected shortfall -- can be learned by minimizing the empirical
score. We parameterize the conditional statistic as a recurrent neural
network acting on the common-noise path, allowing for non-Markovian
dependence, and train it against the elicitable score \cite{coache2023conditionally}. The
same principle yields a pathwise loss for the backward process itself,
which we parameterize as a decoupling field. Two advantages follow.
First, conditional expectations given the common-noise filtration are
computed without nested Monte Carlo simulation, while the solution
remains adapted by construction. Second, because elicitability is not
tied to moments, the framework accommodates mean-field interactions
mediated by general elicitable functionals, such as quantiles or
tail-risk measures, with no change to the algorithm.

We validate the method on three problems of increasing complexity. First, the
systemic-risk interbank borrowing and lending model of
\cite{carmona2013mean}, which 
admits a closed-form solution, serves as a quantitative benchmark and our algorithm recovers the solution accurately. Second, a modification of that model, in which the interaction is mediated by a population quantile rather
than the mean, illustrates the flexibility of the elicitability
framework beyond conditional moments. Finally, we solve a
non-stationary Aiyagari--Bewley--Huggett economic growth model with
endogenous interest rates and common noise \cite{achdou2022income}, demonstrating applicability to mean-field games without closed-form solutions.

\subsection{Related literature}
A survey of numerical methods for BSDEs, including deep learning
approaches, can be found in \cite{chessari2023numerical}. \cite{chassagneux2019numerical} develop a numerical method, which is closest to our scheme, for MV-FBSDEs based on recursive Picard iterations over small time
intervals, computing empirical conditional expectations for $Y$ on each
subinterval. Our method differs in performing each Picard iteration over the whole horizon at once, with the conditional expectation of $Y$ obtained by elicitability rather than by interval-wise regression.

Several recent works solve MV-FBSDEs without common noise with neural networks. \cite{germain2022numerical} approximate $Y$ with deep learning, penalizing the
terminal condition $Y_T$ and computing the mean-field interaction
empirically over sample paths. \cite{zhou2024deep} recast mean-field control
as a characteristic system of ODEs, using score functions and density
estimation to avoid sampling Brownian paths. \cite{han2024learning} first learn
the dependence of the coefficients on the law of the process and then
apply a Deep BSDE solver to the backward component. Our method instead
estimates the law dependence by eliciting the required conditional
statistic directly, which extends naturally to the common-noise setting
and defines a loss over the entire time interval rather than only at
the terminal time.

A complementary line of work avoids neural networks by exploiting
structural assumptions on the measure dependence. Balata, Hur\'e,
Lauri\`ere, Pham, and Pimentel~\cite{Balata19} study a
\emph{polynomial conditional} class of McKean--Vlasov control problems with
common noise, in which the coefficients are affine in the state, the
costs are polynomial, and the law enters only through its first $p$
moments. Under these assumptions, a Markovian embedding tracks the
conditional moments $Y_t^{(k)} = \mathbb{E}[X_t^k \mid \mathcal{F}_t^0]$
as a finite-dimensional system driven by the common noise, reducing the
problem to a standard stochastic control problem solved by
quantization, control randomization, and regress-later Monte Carlo.
This reduction is exact and efficient within its scope, but it is
fundamentally tied to interactions mediated by finitely many moments
and to the polynomial structure of the coefficients. Our approach
removes both restrictions: by eliciting the conditional statistic $S_t$
directly, we accommodate general elicitable functionals and solve the
fully coupled forward--backward system characterizing the mean-field
equilibrium.

More recently, \cite{Hu2025}\footnote{This work appeared contemporaneously with the original version of our paper.} tackle MV-FBSDEs with common noise and use path signatures to encode the common noise. They integrate signature representations with Deep BSDE solvers inside a fictitious-play loop:
at each iteration, the conditional law $\mathcal{L}(\Theta_t \mid \mathcal{F}_t^0)$ is encoded via the truncated (log-)signature of the time-augmented common-noise path $\widehat{W}^0_{[0,t]}$, the distribution-dependent coefficients are fitted by supervised learning, and the resulting conditional FBSDE is solved, with convergence established in the number of fictitious-play iterations up to the supervised-learning error. Like ours, this approach permits general nonlinear dependence on the measure flow. The two methods differ in how the common-noise dependence is captured: rather than reconstructing a finite-dimensional surrogate of the conditional law and learning the measure-embedding maps that act on it, we elicit the statistic $S_t$ required by the coefficients directly, parameterized as a recurrent
network on the common-noise path. This sidesteps the truncation-order
and signature-dimension trade-offs inherent in the signature
representation when the interaction is elicitable, and avoids nested Monte Carlo. We achieve comparable results in our numerical experiments, as shown in Table \ref{tab:mee}. 

Finally, the iterative philosophy underlying our scheme is rooted in
the learning-dynamics literature for mean-field games. Fictitious
play -- in which each agent best-responds to a running time-average of
the previously observed population behaviour -- was introduced in the
mean-field setting by Cardaliaguet and Hadikhanloo \cite{CH15}, who
established convergence for both second-order and first-order MFG
systems whenever the game is potential, i.e., when the running and
terminal couplings derive from potentials on the space of measures. Our
setting differs in that we solve a fully coupled MV-FBSDE rather than a
Hamilton--Jacobi/Fokker--Planck pair, and we do not assume a potential
structure; nevertheless, the averaging mechanism at the heart of
fictitious play directly motivates the soft-update damping
\eqref{eq:soft_updates} that stabilizes our outer Picard iteration. In
this sense, each outer iteration acts as a learning step in which the
measure-flow statistic $S$ and the decoupling field are updated against
a damped average of their previous estimates, in the spirit of the
relaxed fictitious-play schemes discussed in \cite{lauriere2021numerical}.

The remainder of the paper is organized as follows.
Section~\ref{sec:problem} formulates the MV-FBSDE with common noise,
reviews the elicitability framework, and presents the proposed
algorithm in detail. Section~\ref{sec:experiments} reports the
numerical experiments on the systemic-risk model, its
quantile-interaction variant, and the economic growth model.
Section~\ref{sec:conclusion} concludes.

\section{Problem Formulation}\label{sec:problem}

We work on a complete probability space $(\Omega,\mathfrak{F}, \P)$ on which we have two independent, $d$-dimensional Brownian motions $(W_t)_{t \in [0,T]}$ and $(W^0_t)_{t \in [0,T]}$ over a time horizon $T > 0$, where $W^0$  and $W$ represent common and idiosyncratic noise, respectively. We denote the space of square-integrable $\mathcal{A}$--measurable random variables by $\L^{2}[\mathcal{A}]$ and adopt the following notation:
\begin{itemize}[label=$\vartriangleright$]
    \item $\F^0 = (\F^0_t)_{\tT}$, $\F^0_t = \sigma\left( W^0_s, s\leq t \right)$, for the filtration generated by the common noise;

    \item $\F = (\F_t)_{\tT}$, $\F_t = \sigma\left( W_s, s\leq t \right)$, for the filtration generated by the idiosyncratic noise.
\end{itemize}

Our goal is to approximate the solution to the fully-coupled multidimensional MV-FBSDE describing the dynamics of $(X,Y) \in \R^\ell \times \R^m$:
\begin{align}
   \dX_t &= \mu(t, X_t, Y_t, Z_t, Z_t^0, S_t) \,\dt + \sigma(t, X_t) \,\dW_t + \sigma_0(t, X_t) \,\dW^0_t, & X_0 = \xi, 
   \label{eq:forward_sde}
   \\
   \dY_t &= -f(t, X_t, Y_t, Z_t, Z_t^0, S_t) \,\dt + Z_t^\intercal\,\dW_t + {Z_t^0}^\intercal\,\dW^0_t, & Y_T = G(X_T, S_T),\label{eq:backward_sde}
\end{align}
where $\mu: [0,T] \times \R^\ell \times \R^m \times \R^{d \times m} \times \R^{d \times m} \times \R^p \longrightarrow \R^\ell$, $\sigma : [0,T] \times \R^\ell \longrightarrow \R^{\ell \times d}$, $\sigma_0 : [0,T] \times \R^\ell \longrightarrow \R^{\ell \times d}$, $f: [0,T] \times \R^\ell \times \R^m \times \R^{d \times m} \times \R^{d \times m} \times \R^p \longrightarrow \R^m$ and $G : \R^\ell \times \R^p\longrightarrow \R^m$.

The quantity $S_t$ is any elicitable, $p$-dimensional statistic (see \cite{fissler2016higher}) of the state $X_t$, conditioned on the path of the common noise up to time $t$, i.e. we assume there exists a \textit{score function} $\mfS$ such that
\begin{equation}\label{eq:elicit_intro}
   S_t = \argmin_{\mfs \in \L^{2}[\F^0_t]} \E\left[\mfS(\mfs, X_t) \right].
\end{equation}
Equivalently, $S_t$ is an elicitable statistic of the conditional measure $\mu_t = \mathcal{L}(X_t | \mathcal{F}^0_t)$.
For example, we may elicit conditional moments of $X_t$ by choosing $\mfS(\mfs, x) = (\phi(x) - \mfs)^2$ and conditional $\alpha$-quantiles of $X_t$ by choosing $\mfS(\mfs, x) = (1_{\{\mfs \geq x\}} - \alpha)(\mfs - x)$. We provide further details on elicitability in Section \ref{methodology:elicitability} below. This model falls into the setting of FBSDEs in a random environment, \cite[Chapter 1, Vol. II]{carmona2018probabilistic}, in which such environment is given by the elicitable statistic $(S_t)_{t \in [0,T]}$. 

We assume the initial condition $\xi$ is in $\L^2[\G_0]$ and takes values in $\R^\ell$, for a given initial information $\G_0$. Furthermore, the full information, i.e. the filtration generated by both Brownian motions and the initial condition, is denoted by $\G = (\G_t)_{\tT}$, $\G_t = \F_t \vee \F^0_t \vee \sigma(\xi)$. Below, we use  $\|\cdot\|$ to denote the appropriate Euclidean norm. 

We now state the standing assumptions of our work.

\begin{assumption}[Lipschitz Continuity]
\label{asm:lip}
There exist $L, C, L_0 > 0$ such that the following properties hold:
\begin{enumerate}[label=(\roman*),itemsep=0.5em]

\item Linear growth: 
$$\|\mu(t,0,0,0,0,\mfs)\| + \|f(t,0,0,0,0,\mfs)\| + \|\sigma(t,0)\| + \|\sigma_0(t,0)\| \leq C(1 + \|\mfs\|),$$
for all $t \in [0,T]$ and $\mfs \in \R^p;$

\item Lipschitz continuous coefficients: 
\begin{align*}
&\|\mu(t,x,y,z,z^0,\mfs) - \mu(t,x',y',z',z'^0,\mfs)\| + \|f(t,x,y,z,z^0,\mfs) - f(t,x',y',z',z'^0,\mfs)\| \\
&+ \|\sigma(t,x) - \sigma(t,x')\| + \|\sigma_0(t,x) - \sigma_0(t,x')\| + \|G(x,\mfs) - G(x',\mfs)\| \\
&\leq L(\|x - x'\| + \|y - y'\| + \|z - z'\| + \|z^0 - z'^0\|),    
\end{align*}
for all $t \in [0,T]$ $x, x' \in \R^\ell$, $y,y' \in \R^m$, $z,z^0,z',z'^0 \in \R^{d \times m}$, and $\mfs \in \R^p;$

\item Lipschitz continuous $Y$: 

$\|Y_t^{t,x} - Y_t^{t,x'}\| \leq L_0 \|x - x'\|$, for any $t \in [0,T]$ and $x, x' \in \R^\ell$, where $(X_s^{t,x},Y_s^{t,x},$ $Z_s^{t,x},Z_s^{0^{t,x}})_{s \in [t,T]}$ is a solution for \eqref{eq:forward_sde}--\eqref{eq:backward_sde} for deterministic initial condition $x$ at time $t$.

\end{enumerate}    
\end{assumption}

Under Assumptions~\ref{asm:lip}, by \cite[Theorem 1.45, Proposition 1.52]{carmona2018probabilistic}, there exists a unique solution $(X,Y,Z,Z^0)_{t \in [0,T]}$ for \eqref{eq:forward_sde}--\eqref{eq:backward_sde} such that
\begin{align}
\E \left[\sup_{t \in [0,T]} \Big\{\|X_t\|^2 + \|Y_t\|^2\Big\} 
+ \int_0^T \Big(\|Z_t\|^2 + \|Z^0_t\|^2\Big) dt \right] < \infty.
\end{align}
Moreover, the backward process $Y$ admits a representation of the form
\begin{align}\label{eq:decoupling_field}
Y_t = U(t,X_t,(W^0_s)_{s \leq t}),
\end{align} 
for some function $U : [0,T] \times \R^\ell \times C([0,T];\R^d) \rightarrow \R^m$ known as the \textit{decoupling field} \cite[Proposition 1.50, Remark 1.49]{carmona2018probabilistic}. 

\begin{remark}
The decoupling field is classically represented as a function of $(t,X_t,\mathcal{L}(X_t \mid \mathcal{F}^0_t))$. While this formulation provides additional structure, it requires the estimation of the path-dependent conditional law $\mathcal{L}(X_t \mid \mathcal{F}^0_t)$. In contrast, representing the decoupling field as a function of $(t,X_t,(W^0_s)_{s\le t})$ allows the algorithm to learn directly the dependence of $Y$ on the common-noise path, avoiding the estimation of the conditional law whenever possible. Under additional assumptions, notably in the linear--quadratic case, this framework also yields a Markovian representation of the decoupling field in terms of $(t,X_t,S_t)$.
\end{remark}

Integrating \eqref{eq:backward_sde}, we see that $Y$ admits the following fixed-point representation
\begin{align}\label{eq:y_intro}
Y_t = \E\left[\left. G(X_T, S_T) + \int_t^T f(u,X_u, Y_u, Z_u, Z_u^0, S_u)\du \ \right| \ \G_t \right].
\end{align}
We can rewrite this fixed-point representation using the elicitability of the mean (or equivalently the $L^2$-projection representation for conditional expectations) via a quadratic score function:
\begin{equation}
\label{methodology:backward_elicitability}
    Y_t = \argmin_{\mfY \in \L^{2}[\G_t]} \E\left[ \left(\mfY - G(X_T, S_T) - \int_t^T f(u,X_u,Y_u,Z_u,Z_u^0,S_u)\, \,\du\right)^2  \right].
\end{equation}
Moreover, by the decoupling field representation, the minimizer in \eqref{methodology:backward_elicitability} must have the form $\mfY^*=U(t,X_t,(W^0_s)_{s \leq t})$. 
The processes $Z$ and $Z^0$ admit a representation using the It\^o's formula along a flow of conditional measures \cite[Vol II, Theorem 4.17]{carmona2018probabilistic}. Using that representation, however, presents challenges for numerical implementation. Instead, in Section \ref{sec:approx_ZandZ0}, we use another (approximate) representation that elicits them.

\subsection{Proposed method}

We next provide an overview of our methodology, and provide further details in Section \ref{sec:detailed-description}. Our methodology proceeds iteratively. We fix a time discretization of $[0,T]$, denoted by $\T = \{t_0,\dots,t_N\}$, where $0=t_0<\cdots < t_N=T$. Given samples of $(W_t, W_t^0)_{\tT}$ and initialization -- which represents the initial guess for the solution to the MV-FBSDE system -- $(X^0_t, Y^0_t, Z^0_t, Z^{0,0}_t, S^0_t)_{\tT}$, we update, at iteration $k \in \mathbb{N}$, each entry of $(X^k_t, Y^k_t, Z^k_t, Z^{0,k}_t, S^k_t)_{\tT}$ sequentially as follows:

\begin{enumerate}[label=(\roman*),itemsep=0.5em]

\item First, we solve the forward SDE \eqref{eq:forward_sde} for $(X^{k+1}_t)_{\tT}$,  through Picard iterations, with $(W_t, W_t^0)_{\tT}$ and $(Y^k_t, Z^k_t, Z^{0,k}_t, S^k_t)_{\tT}$
held fixed. We denote the $n$-th Picard iteration by $X_t^{k+1,n}$, set $X_t^{k+1,0}=X_t^{k}$, and repeat this inner Picard iteration until we achieve the desired convergence error in the $L^2$--norm, say at iteration $\mfn_{k+1}$. We then set $X^{k+1}=X^{k+1,\mfn_{k+1}}$ resulting in the $(k+1)^{th}$--outer Picard iteration of the process $X$.

\item Next, using $(W_t, W_t^0)_{\tT}$ and $(X^{k+1})_{\tT}$, we compute $S^{k+1}_t$ using elicitability, i.e. we solve the minimization problem \eqref{eq:elicit_intro} using $X^{k+1}_t$. We approximate $\mfs \in \L^{2}[\F^0_t]$ as a recurrent neural network $\mfs(t,(W^0_s)_{s\leq t})$, for $\tT$.

\item Finally, with $(W_t, W_t^0)_{\tT}$, $(X^{k+1}_t, S^{k+1}_t)_{\tT}$ and $(Y^k_t,Z^k_t, Z^{0,k}_t)_{\tT}$ fixed, we solve for $(Y^{k+1}_t, Z^{k+1}_t, Z^{0,k+1}_t)_\tT$ by calculating the conditional expectation for $Y^{k+1}$ through elicitability using \eqref{methodology:backward_elicitability}. We parameterize $Y^{k+1}_t \in \L^2[\G_t]$  as $U(t,X^{k+1}_t,(W_s^0)_{s \leq t})$ (the \textit{decoupling field}) and approximate $U$ as the output of a neural network. 
Finally, we obtain $Z^{k+1}_t$ and $Z^{0,k+1}_t$ using elicitability as described in Section \ref{sec:approx_ZandZ0}.

\end{enumerate}

These steps are summarized in Algorithm \ref{methodology:overall_algorithm} and the code is available at \url{https://github.com/fjpAntunes/mean-field-tools/tree/main/mean_field_tools/deep_bsde}.
\begin{algorithm}[t]
\SetAlgoLined
\KwIn{Drift $\mu(u,x,y,z,z^0,s)$; diffusion coefficients $\sigma(u,x)$ and $\sigma_0(u,x)$; Initial condition $X_0$; Terminal condition $g(x,s)$; driver function $f(u,x,y,z,z^0,s)$; Time discretization $\mathcal{T}$; Samples of $(W_t, W_t^0)_\tT$;  Initial $(Y^0_t, Z^0_t, Z^{0,0}_t, S^0_t)_{\tT}$; Number of Picard iterations $K$, Resampling frequency $R$.}
\KwOut{Samples of $(X^K_t, Y^K_t, Z^K_t, Z^{0,K}_t, S^K_t )_\tT$ providing an approximate solution to the system \eqref{eq:forward_sde}-\eqref{eq:backward_sde}; Trained neural networks $S_\theta$, $U_\theta$, $V_\theta$, $V^0_\theta$.}

\For{$k = 1$ to $K $}{

    Given $(X^k_t, Y^k_t, Z^k_t, Z^{0,k}_t, S^k_t )_\tT$, use Algorithm \ref{methodology:numerical_forward_algorithm} to sample $(X^{k+1}_t)_\tT$ \;

    Given $(X^{k+1}_t)_\tT$, use Algorithm \ref{methodology:law_elicitability_algorithm} to sample $(S^{k+1}_t)_\tT$ \;

    Given $(X^{k+1}_t, Y^k_t, Z^k_t, Z^{0,k}_t, S^{k+1}_t )_\tT$, use Algorithm \ref{methodology:backward_elicitability_algorithm} to sample $(Y^{k+1}_t, Z^{k+1}_t)_\tT$ \;

    Given $(X^{k+1}_t, Y^{k+1}_t, Z^{k+1}_t, Z^{0,k}_t, S^{k+1}_t )_\tT$, use Algorithm \ref{methodology:backward_vol_elicitability_algorithm} to sample $(Z^{0,k+1}_t)_\tT$.

    Every $R$ iterations, regenerate Brownian motions and use Algorithm \ref{methodology:sampling_after_training} to resample paths.

}

\Return{Samples of $(X^K_t, Y^K_t, Z^K_t, Z^{0,K}_t, S^K_t )_\tT$ }

\caption{Picard Iterations and Elicitability for MV-FBSDE}
\label{methodology:overall_algorithm}
\end{algorithm}

To improve convergence and stability of the algorithm, we dampen the full iteration by performing `soft-updates' of the processes as follows
\begin{align}\label{eq:soft_updates}
   \Psi^{k+1} &= \delta \,\Psi^k + (1-\delta)\,{\widehat \Psi}^{k+1},
\end{align}
where $\Psi$ is $X$, $Y$, $Z$ and $Z^0$ and $0<1-\delta\ll 1$ is a damping coefficient and hatted variables represent the updates computed from the steps above.

We continue the main outer iteration until a maximum number of iterations is reached, or the $\L^2$--norm between samples of successive iterations of $X$, $Y$, $Z$ and $Z^0$ fall below a specified tolerance.

\subsection{Detailed description}
\label{sec:detailed-description}

We next provide a detailed description of our methodology. As outlined above, our method consists in iterating between fixing the backward SDE and solving the forward SDE by Picard iterations, and fixing the forward SDE and solving the BSDE by elicitability. Each step of this process consists in a Picard iteration for the whole MV-FBSDE system. The MV aspect requires estimating $S_t$, which is achieved by elicitability as well.

We remind the reader that the outer-Picard iteration of the backward SDE is indexed by $k$ and, for each $k$, the inner-Picard iteration of the forward SDE is indexed by $n$. 

\subsubsection{Solving the forward SDE by Picard iterations}\label{methodology:forward} 

Suppose we are at the $k^{th}$ outer iteration, i.e., $(X^k, Y^k, Z^k, Z^{0,k}, S^k)$ are fixed. We define $X^{k+1,n}$ to the be $n^{th}$--inner Picard update of the forward process. To this end, set $X^{k+1,0}=X^{k}$, and given $X^{k+1,n}$, we obtain $X^{k+1,n+1}$ through
\begin{align*}
   X^{k+1,n+1}_t = X_0 &+ \int_0^t \mu(u, X^{k+1,n}_u, Y^k_u, Z^k_u, Z^{0,k}_u, S^k_u) \,\du \\
   &+ \int_0^t \sigma(u, X^{k+1,n}_u) \,\dW_u + \int_0^t \sigma_0(u, X^{k+1,n}_u) \,\dW_u^0\,,
\end{align*}%
by numerically evaluating the right-hand side using Euler-Maruyama method with time discretization $\mathcal{T}$ until $\left\| X^{k+1,n+1} - X^{k+1,n} \right\|_{\L^2(\Omega \times [0,T])}$ is below a specified tolerance; see Algorithm \ref{methodology:numerical_forward_algorithm} for a pseudo-code, where we have suppressed the outer iteration counter, $k$, for simplicity. The counter $i$ denotes the Monte arlo samples.

\begin{algorithm}
\SetAlgoLined
\KwIn{Drift $\mu(u,x,y,z,z^0,s)$; diffusion coefficients $\sigma(u,x)$ and $\sigma_0(u,x)$; Initial condition $X_0$; Time discretization $\mathcal{T}$; Tolerance parameter $\varepsilon$; $M$ samples of $(W_t, W_t^0)_\tT$; samples of $(Y_t,Z_t,Z_t^0,S_t)_\tT$.}
\KwOut{$M$ samples of $(X_t)_\tT$ at the next Picard iteration.}

Initialize $X^0_t \equiv X_0$ 

$n \gets 0$\;
\While{$\mfe \geq \varepsilon$}{
    Using Euler-Maruyama in $\mathcal{T}$, approximate $M$ samples of $$X^{n+1}_t = X_0 + \int_0^t \mu(u, X^n_u, Y_u, Z_u, Z^0_u, S_u) \, \,\du + \int_0^t \sigma(u, X^n_u) \,\dW_u + \int_0^t \sigma_0(u, X^n_u) \,\dW^0_u;$$
    Compute error $$\mfe = \frac{1}{MN}\sum_{j=1}^N \sum_{i=1}^M ||X^{n+1,(i)}_{t_j} - X^{n,(i)}_{t_j}||^2;$$
    $n \gets n + 1$\;
}
\Return{$X^{n+1}$}
\caption{Picard iteration of forward SDE with fixed $(Y,Z,Z^0,m)$}
\label{methodology:numerical_forward_algorithm}
\end{algorithm}

\subsubsection{Estimating $S$ through elicitability}\label{methodology:elicitability}

Elicitability provides a framework for computing statistics through minimization of scoring functions \cite{fissler2016higher}. For example, given a random variable $X$ taking values in $\mathbb{R}$, and a $\sigma$-algebra $\mathcal{G}$, the conditional expectation $\E\left[X \mid \mathcal{G}\right]$ can be characterized as the minimizer of the $L^2$ loss:
\begin{equation}
    \E\left[X|\mathcal{G}\right] = \argmin_{Y \in \L^2(\mathcal{G})} \E\left[\big(X-Y\big)^2\right].
\end{equation}
This characterization transforms the problem of computing conditional expectations into an minimization problem over $\mathcal{G}$-measurable random variables with second moment. 

At the $k^{th}$ outer iteration, we have samples from $(X^k, Y^k, Z^k, Z^{0,k}, S^k)_\tT$ and in the previous section, we have obtained samples from $(X^{k+1})_\tT$. We next estimate $S^{k+1}$.

We begin by analyzing the cases where the dependency on $S$ is through some conditional moment $S_t = \E[\phi(X_t) \mid \F_t^0]$. When there is no common noise, $S^{k+1}_t$ can be estimated by calculating the empirical average over samples of $\phi(X^{k+1}_t)$. In the case where there is common noise this would be numerically prohibited, as we need to condition on the state of common noise. Instead, we use elicitability and rewrite the conditional expectation as a minimization problem
\begin{equation}\label{methodology:common_noise_conditional_mean_elicitability}
    S^{k+1}_t = \argmin_{\mfs \in \L^{2}[\F^0_t]} \E \left[ \left(\phi(X_t^{k+1}) - \mfs \right)^2 \right].
\end{equation}
More generally, if $S_t$ is any elicitable statistic for a score function $\mfS$ over the conditional probability measure of $X_t$ given $\F^0_t$, we may write
\begin{equation}\label{eq:elicit}
   S_t^{k+1} = \argmin_{\mfs \in \L^{2}[\F^0_t]} \E\left[\mfS(X_t^{k+1}, \mfs)\right].
\end{equation}
For example, we may elicit the (conditional) $\alpha$-quantile by using the score function
\begin{equation}\label{eq:alpha_quantile}
    \mfS(x,\mfs) = \left(1_{\{x \geq \mfs\}} - \alpha\right)\,\left(x - \mfs\right).
\end{equation}
On the other hand, $\alpha$-expected shortfall is not elicitable on its own, but it is $2$-elicitable, meaning the $\alpha$-quantile and $\alpha$-expected shortfall are jointly elicitable. A brief exposition of elicitability can be found in \cite{pesenti2024risk}.

In general, the solution of the optimization problem in \eqref{eq:elicit} may not be  Markovian in $W^0$. Therefore, we  consider a \textit{recurrent} neural network (RNN) to the parametrize the function $S$, and find an approximate solution for the minimization problem \eqref{methodology:common_noise_conditional_mean_elicitability} through stochastic gradient descent. See Algorithm \ref{methodology:law_elicitability_algorithm} for a pseudo-code, where we suppress the outer iteration counter, $k$, for simplicity.

\begin{algorithm}[H]
\SetAlgoLined
\KwIn{Score function $\mfS$ to elicit $S$; $M$ samples of $(W_t, W_t^0)_\tT$; $M$ samples of $(X_t)_\tT$; Batch size $I$; number of iterations $E$.}
\KwOut{$M$ samples of $(S_t)_\tT$ with respect to the input samples of $(X_t)_\tT$ and $(W_t^0)_\tT$; Trained recurrent neural network $S_\theta$.}

Initialize recurrent neural network $S_\theta$ with random weights $\theta$\;

\For{iter = 1 to $E$}{

        sample $I$ indexes $\{m_1,\ldots,m_I\}$ from $\{1,\ldots,M\}$ with replacement.

        Compute loss: \\
        $\displaystyle L(\theta) = \frac{1}{IN} \sum_{\tT} \sum_{i=1}^I \mfS(X_t^{(m_i)}, S_\theta(t,(W_s^{0(m_i)})_{u \leq t, u \in \mathcal{T}}))$\;
        Update $\theta$ using Adam\;

}
\Return{Trained recurrent neural network $S_\theta$}
\caption{Estimating $S$ through elicitability for fixed $(X,Y,Z, Z^0)$}
\label{methodology:law_elicitability_algorithm}
\end{algorithm}

\subsubsection{Solving the backward SDE through elicitability}\label{methodology:backward}

Given $(X^k, Y^k, Z^k, Z^{0,k}, S^k)$, from the previous two steps have estimated $X^{k+1}$ and $S^{k+1}$, we next update $Y$. Based on the representation \eqref{methodology:backward_elicitability}, we consider the following minimization problem:
\small
\begin{equation}\label{eq:Y_minimization}
   Y^{k+1}_t = \argmin_{\mfY \in \L^2[\G_t]} \E\left[\left(\mfY - \left( G(X^{k+1}_T, S^{k+1}_T) + \int_t^T f\big(u,X^{k+1}_u,Y^k_u,Z^k_u, Z^{0,k}_u,S^{k+1}_u\big) \,\du\right)\right)^2\right].
\end{equation}
\normalsize
The structure given by the decoupling field in Equation \eqref{eq:decoupling_field} implies that we may parameterize the backward process $Y$ using a neural network that takes $(t,X_t,S_t)$ as inputs. 

Our numerical method consists in finding an approximate solution for the minimization problem
\eqref{eq:Y_minimization}. Denote the numerical solution of this minimization procedure by $U_\theta$. Hence, we set $Y_t^{k+1} = U_\theta(t,X^{k+1}_t,(W_u^0)_{u \leq t})$. To improve convergence rates, we calculate the loss with time dependent weights $p_t$, using a higher weight to the terminal condition - typically, $p_T = \frac{N}{2}$ and $p_t = 1$ for $t < T$. We provide a summary of this part of the method in Algorithm \ref{methodology:backward_elicitability_algorithm}, where we have suppressed the counter $k$ for simplicity and we  are using $\theta$ to represent the generic neural network parameters throughout the paper.
\begin{algorithm}
\SetAlgoLined
\KwIn{Terminal condition $g$; driver function $f(u,x,y,z,z^0,s)$; Time discretization $\mathcal{T}$;
weights $p_t$ and sum of weights $P = \sum_{t \in \mathcal{T}} p_t$;
$M$ samples of $(W_t, W_t^0)_\tT$; $M$ samples of $(X_t,S_t)_\tT$ at the next Picard iteration; $M$ samples $(Y_t, Z_t,Z_t^0)_\tT$ at the current Picard iteration; Batch size $I$; number of iterations $E$.}
\KwOut{$M$ samples of $(Y_t, Z_t)_\tT$ at the next Picard iteration; Trained neural network $U_\theta$.}

Initialize neural network $U_\theta$ with random weights $\theta$\;

\For{iter = 1 to $E$}{

            sample $I$ indexes $\{m_1,\ldots,m_I\}$ from $\{1,\ldots,M\}$ with replacement.

            Set $f_u = f(u,X_u,Y_u,Z_u, Z^0_u,S_u)$, for $u \in \mathcal{T}$ \;
    Compute the loss 
     $$\hspace*{-1cm}\mathcal{L}(\theta) = \frac{1}{IP} \sum_{\tT} \sum_{i=1}^I  p_t\left(U_\theta(t,X_t^{(m_i)},(W_s^{0(m_i)})_{u \leq t, u \in \mathcal{T}}) - G(X_T^{(m_i)}, S_T^{(m_i)}) - \sum_{u \in \T, u \geq t} f_u^{(m_i)}\, \,\Delta t\right)^2;$$
            Update $\theta$ using Adam\;

    }

\Return{Trained neural network $u_\theta$}

\caption{Estimating $Y$ through elicitability with fixed $(X,Z,Z^0,m)$}
\label{methodology:backward_elicitability_algorithm}
\end{algorithm}

\subsubsection{Approximating $Z$ and $Z_0$}\label{sec:approx_ZandZ0}
We consider the following discretization of the backward dynamics of $Y$ \eqref{eq:backward_sde} in the discretized times $\mathcal{T}$:
\begin{equation}
\Delta Y_j^{k+1} \coloneqq Y_{t_{j+1}}^{k+1} - Y_{t_j}^{k+1} \approx - f_j^{k+1} \, (t_{j+1} - t_j) + Z_{t_j}^{k+1\intercal} \Delta W_j + Z_{t_j}^{0,k\,\intercal} \Delta W^0_j,
\end{equation}
where, for $j=0,\ldots,N-1$, $f_j^{k+1} = f(t_j,X_{t_j}^{k+1},Y_{t_j}^{k+1},Z_{t_j}^{k+1}, Z_{t_j}^{0,k},S_{t_j}^{k+1})$.
We may then update $Z^0_{t_j}$ by computing
\begin{align}\label{eq:Z0}
  Z_{t_j}^{0,k+1}  &= \mathbb{E}\left[ \left.\left(\frac{ \Delta Y_j^{k+1} }{\Delta t} + f_j^{k+1} \right)\Delta W^0_j \right|\, \mathcal{G}_{t_j}\,\right].
\end{align}
Thus, as we have samples of $(Y_t^{k+1})_\tT$ and $(f_t^{k+1})_\tT$ from the previous steps, we may elicit $Z^{0,k+1}$ by
\begin{equation}
Z_{t_i}^{0,k+1}  = \argmin_{ \mathfrak{Z} \in \L^{2}[\G_{t_i}]} \E\left[ \left(\mathfrak{Z} -  \left(\frac{ \Delta Y_i^{k+1} }{\Delta t} + f_i^{k+1} \right) \Delta W^0_i\right)^2 \right].
\end{equation}
Additionally, $Z_{t_i}^{k+1}$ can be obtained in the same way as above, replacing $\Delta W_i^0$ with $\Delta W_i$.
We provide a summary of this part of the method in Algorithm \ref{methodology:backward_vol_elicitability_algorithm}. Since they may depend fully on the path of the common noise, we parametrize them as a recurrent neural network that takes $(t,X_t,(W_s^0)_{s \leq t})$ as inputs, to encode this path-dependence.

\begin{algorithm}[H]
\SetAlgoLined
\KwIn{Driver function $f(u,x,y,z,z^0,s)$; Time discretization $\mathcal{T}$; $M$ samples of $(W_t, W_t^0)_\tT$; $M$ samples of $(X_t,S_t,Y_t, Z_t)_\tT$ at the next Picard iteration; $M$ samples of $(Z_t^0)_\tT$ at the current Picard iteration; Batch size $I$; number of iterations $E$.}
\KwOut{$M$ samples of $(Z_t, Z^0_t)_\tT$ at the next Picard iteration; Trained recurrent neural networks $v_\theta$ and $v_\theta^0$.}

Initialize recurrent neural networks $V_\theta$ and $V_\theta^0$ with random weights $\theta$\;

\For{iter = 1 to $E$}{

            sample $I$ indexes $\{m_1,\ldots,m_I\}$ from $\{1,\ldots,M\}$ with replacement.

            Set $f_u = f(u,X_u,Y_u,Z_u,Z^0_u,S_u)$ \;
            Calculate $\Delta Y^{(m_i)}_t \coloneqq Y^{(m_i)}_{t + \Delta t} - Y^{(m_i)}_t$
    Compute the loss $$\mathcal{L}(\theta) = \frac{1}{IN} \sum_{\tT\setminus \{T\}} \sum_{i=1}^I \left(V_\theta(t,X_t^{(m_i)},(W_s^{0(m_i)})_{s \leq t, s \in \mathcal{T}}) - \left( \frac{\Delta Y^{(m_i)}_t}{\Delta t} + f_t^{(m_i)} \right) \,\Delta W_t^{(m_i)}\right)^2;$$
    Similar loss for $V_\theta^0$ \;
            Update $\theta$ using Adam\;

}

\Return{Trained recurrent neural networks $V_\theta$ and $V_\theta^0$}

\caption{Estimating $Z$ and $Z^0$ through elicitability with fixed $(X,Y,m)$}
\label{methodology:backward_vol_elicitability_algorithm}
\end{algorithm}

\subsection{Sampling after training} 

After completing Algorithm \ref{methodology:overall_algorithm} to the desired accuracy, we have approximate samples of the processes $(X,Y,Z,Z^0,S)$ over the discretized times $\mathcal{T}$. Moreover, as a product of the algorithm, we also have neural networks to approximately sample $Y$, $Z$, $Z^0$, given samples of $(W,W^0)$ \textit{and} $X$. Therefore, samples of the full set of processes $(X,Y,Z,Z^0,S)$ after training may be obtained by iterating with the trained neural networks, as described in Algorithm \ref{methodology:sampling_after_training}.

\begin{algorithm}[H]
\SetAlgoLined
\KwIn{Trained neural networks $U_\theta, V_\theta, V^0_\theta, S_\theta$ ,drift $\mu(u,x,y,z,z^0,s)$; diffusion coefficients $\sigma(u,x)$ and $\sigma_0(u,x)$; Initial condition $X_0$; Time discretization $\mathcal{T}$, Samples of $(W_t,W_t^0)_\tT$, Initial sample of $(X^{0}_t)$, number of iterations $N$.}
\KwOut{Approximate samples of $(X_t, Y_t, Z_t, Z^{0}_t, S_t )_\tT$}

Set $S_t^\theta = S_\theta(t,(W_s^{0})_{s \leq t})$.

Set 
$\begin{cases}
Y^\theta_t = U_\theta(t, X^{0}_t, (W_s^{0})_{s \leq t}),\\
Z^\theta_t = V_\theta(t, X^{0}_t, (W_s^{0})_{s \leq t}), \\
Z^{0,\theta}_t = V^0_\theta(t, X^{0}_t, (W_s^{0})_{s \leq t}).    
\end{cases}$

\For{$i = 1$  to $N$}{
Use Algorithm~\ref{methodology:numerical_forward_algorithm} to generate samples $(X^{i+ 1}_t)_\tT$

Set
$\begin{cases}
Y^\theta_t = U_\theta(t, X^{i}_t,  (W_s^{0})_{s \leq t}),\\
Z^\theta_t = V_\theta(t, X^{i}_t,  (W_s^{0})_{s \leq t}), \\
Z^{0,\theta}_t = V^0_\theta(t, X^{i}_t,  (W_s^{0})_{s \leq t}).    
\end{cases}$

}
\tiny
\Return{Samples of $\left(X^{N}_t, 
U_\theta(t, X^{N}_t,(W_s^{0})_{s \leq t}),
V_\theta(t, X^{N}_t,  (W_s^{0})_{s \leq t}),
V^0_\theta(t, X^{N}_t, (W_s^{0})_{s \leq t}),
S_\theta(t,(W_s^{0})_{s \leq t})
\right)_\tT$ }

\caption{Approximate samples of MV-FBSDE after training}
\label{methodology:sampling_after_training}
\end{algorithm}

\section{Error Bound Analysis}\label{sec:convergence}

In this section, we analyze the outer Picard iteration of Algorithm~\ref{methodology:overall_algorithm}. We remind the reader of our standing Assumption~\ref{asm:lip} 
and denote the unique solution of \eqref{eq:forward_sde}--\eqref{eq:backward_sde} by $(X^\star, Y^\star, Z^\star, Z^{0,\star},$ $S^\star)$.

We require the following additional assumptions to derive the error bound analysis:

\begin{assumption}[Lipschitz continuity of coefficients in $s$]\label{asm:lip_s}
The coefficients $\mu,f$ and $G$ are Lipschitz in the measure-flow statistic: there exists $L_s > 0$ such that for all $t, x, y, z, z^0$:
\begin{align*}
\|\mu(t,x,y,z,z^0,\mfs_1) - \mu(t,x,y,z,z^0,\mfs_2)\| &+ \|f(t,x,y,z,z^0,\mfs_1) - f(t,x,y,z,z^0,\mfs_2)\| \\
& + \|G(x,\mfs_1) - G(x,\mfs_2)\| \leq L_s \, \|\mfs_1 - \mfs_2\|,    
\end{align*}
for all $\mfs_1, \mfs_2 \in \R^p$.
\end{assumption}

This condition is separate from Assumption~\ref{asm:lip}(ii), which controls Lipschitz dependence on $(x,y,z,z^0)$ with~$s$ held fixed.  Assumption~\ref{asm:lip_s} governs how errors in the measure-flow approximation propagate into the backward step.  It is satisfied in all three examples of Section~\ref{sec:experiments}: the systemic risk model (where $f$ and $G$ are affine in~$s$), the quantile interaction model, and the Aiyagari--Bewley--Huggett growth model (where the interest rate and wage depend Lipschitz-continuously on the aggregate capital~$\bar k$).

Moreover, let
\(
\mathcal{S}:\mathcal{P}_2(\mathbb{R}^\ell)\to\mathbb{R}^{p}
\)
denote the measure-flow elicitable statistic, meaning
\(
S_t=\mathcal{S}(\mu_t)\), with \(
\mu_t=\mathcal{L}(X_t\mid\mathcal{F}_t^0).
\)

\begin{assumption}[Lipschitz continuity of the statistic]\label{asm:lip_statistic}
The map $\mathcal{S}$ is Lipschitz continuous with respect to the $2$-Wasserstein distance: there exists $L_{\mathcal{S}}>0$ such that, for all
$\mu,\nu\in\mathcal{P}_2(\mathbb{R}^\ell)$,
\begin{equation}
    \|\mathcal{S}(\mu)-\mathcal{S}(\nu)\|
    \leq L_{\mathcal{S}}\,W_2(\mu,\nu).
    \label{eqn:lipschitz-statistic}
\end{equation}
\end{assumption}
Combined with Assumption \ref{asm:lip_s}, we have
\begin{equation}\label{eq:drift_wass_lipschitz}  
\|f(\cdot, \mathcal{S}(\mu))-f(\cdot, \mathcal{S}(\nu))\|\le L_s \|\mathcal{S}(\mu)-\mathcal{S}(\nu)\| \le L_m\,W_2(\mu,\nu),
\end{equation}
where $L_m = L_s\,L_{\mathcal{S}}$. The same is true for $\mu$ and $G$. Note that Assumption~\ref{asm:lip_statistic} is not satisfied by all three examples: quantiles are not Lipschitz continuous with respect to the Wasserstein distance in general. We consider their case separately in Subsection~\ref{ssec:quantile_bound}.

Following~\cite{chassagneux2019numerical}, we start our analysis by isolating the outer Picard iteration from all
other sources of error: we assume an \emph{idealized solver} in which the
inner forward Picard loop (Algorithm~\ref{methodology:numerical_forward_algorithm}) is run to exact convergence and the
elicitation steps (Algorithms~\ref{methodology:law_elicitability_algorithm}, \ref{methodology:backward_elicitability_algorithm}, \ref{methodology:backward_vol_elicitability_algorithm}) return the exact minimizers, so that
$S^{k}$ is the exact conditional statistic of $X^{k}$ and the backward step
returns the exact decoupling field of \eqref{eq:backward_sde} with the measure-flow
argument frozen at $S^{k}$. The only approximation is then the termination
of the outer loop after $K$ iterations. This idealization is dropped in Section~\ref{sec:inexact}, where we generalize the error bounds to include the approximation errors from these neural networks. 

The key idea of our analysis is partitioning the time interval $[0,T]$ into smaller sub-intervals. Fix the equidistant partition $0=t_0<\dots<t_N=T$ with
\[
  h := \tfrac{T}{N}, \qquad I_i := [t_i,t_{i+1}], \quad i=0,\dots,N-1 .
\]
For boundary data $(\zeta,\eta)$, let $\mathcal P_i[\zeta,\eta]$ denote the
\emph{sub-MV-FBSDE} on $I_i$, i.e. \eqref{eq:forward_sde}--\eqref{eq:backward_sde} restricted to
$t\in I_i$, with $X_{t_i}=\zeta$, $Y_{t_{i+1}}=\eta$, the same coefficients,
and $S_t$ elicited from the conditional law of the sub-problem's own
forward solution. 
We write 
\[
  \mathcal E^k_i := \E\!\Big[\sup_{t\in I_i}\norm{ X^k_t - X^\star_t}^2\Big],
  \qquad i = 0,\dots,N-1 .
\]
We also set $\Theta_t:=(Y_t,Z_t,Z^0_t)$ for the backward triple, and define
\[
  \B^k_i
  := \E\!\left[\int_{I_i}\!\norm{\Theta^k_t - \Theta^\star_t}^2\,{{\rm d} t} \right],
  \quad  \B^k \coloneqq \E\!\left[\int_0^T\!\norm{\Theta^k_t - \Theta^\star_t}^2\,{{\rm d} t} \right] = \sum_{i=0}^{N-1} \B_i^k.
\]

\subsection{Local error bound}\label{ssec:local_bound}
The estimate in the proof of Theorem~1.45 in
\cite{carmona2018probabilistic} provides, with the common constant $\Gamma \geq 1$, the following forward and
backward inequalities: 
\begin{align}
  \E\Big[\sup_{t\in I_i}\norm{X_t-X'_t}^2\Big]
  &\le \Gamma\,\E\left[\norm{X_{t_i}-X'_{t_i}}^2\right] 
  + \Gamma h\,\E\! \left[\int_{I_i}\!\norm{\Theta_t-\Theta'_t}^2\,{{\rm d} t} \right] \notag\\
  &\quad{}
  + \Gamma\,\E\!\left[ \int_{I_i}\!
     \big\|\mu(t,X_t,\Theta_t,S_t)
  -\mu(t,X_t,\Theta_t,S'_t)\big\|^2\,{{\rm d} t} \right],
     \label{eq:apriori-fwd}\\[3pt]
  \E\! \left[\int_{I_i}\!\norm{\Theta_t-\Theta'_t}^2\,{{\rm d} t} \right]
  &\le \Gamma\,\E \left[\norm{Y_{t_{i+1}}-Y'_{t_{i+1}}}^2 \right] \notag\\
  &\quad{}+ \Gamma\,\E\!\left[ \int_{I_i}\!
     \big\|f(t,X_t,\Theta_t,S_t)-f(t,X'_t,\Theta_t,S'_t)\big\|^2\,{{\rm d} t} \right] .
     \label{eq:apriori-bwd}
\end{align}

\begin{lemma}[Local error bound up to boundary mismatches]
\label{lem:decay}
Under the standing idealization and Assumptions~\ref{asm:lip},~\ref{asm:lip_s} and~\ref{asm:lip_statistic}, let $\Gamma=\Gamma(L)\ge1$
be the stability constant of~\eqref{eq:apriori-fwd}-\eqref{eq:apriori-bwd}, which depends only on the Lipschitz constant $L$ of Assumption~\ref{asm:lip}. For every $i\in\{0,\dots,N-1\}$ and every
$k\ge1$,
\begin{equation}
  \mathcal E^k_i
  \;\le\; \lambda(h)\,\mathcal E^{k-1}_i
        \;+\; \Gamma\,\delta^k_i
        \;+\; \Gamma^2 h\,\tau^{k-1}_i,
\label{eq:onesweep}
\end{equation}
where
\begin{equation}
      \lambda(h)\coloneqq\Gamma h \left(\,2\Gamma h (L^2 + L_m^2) + L_m^2\,\right)
\end{equation}
and the two \emph{boundary mismatches} are
\[
  \delta^k_i:=\E\norm{X^k_{t_i} - X^\star_{t_i}}^2 \quad(\text{initial, at }t_i),
  \qquad
  \tau^k_i:=\E\norm{ Y^k_{t_{i+1}} - Y^\star_{t_{i+1}}}^2 \quad(\text{terminal, at }t_{i+1}).
\]
In particular, we may choose $N$ large enough that $\lambda(h)<1$; then on each subinterval the iterations contract \emph{up to} the two
boundary sources $\delta^k_i,\tau^{k-1}_i$.
\end{lemma}

\begin{proof}
Fix $i\in\{0,\dots,N-1\}$ and $k\ge1$. The bound is the per-subinterval
instantiation of the two a priori stability estimates underlying
\cite[Thm.~1.45]{carmona2018probabilistic}. The constant $\Gamma=\Gamma(L)\ge1$ above comes from the two estimates: it depends only on the
joint Lipschitz constant $L$ of $(\mu,\sigma,\sigma^0,f,G)$, and not on $h$, $N$, or $k$. We write
$\mu^j_t:=\mathcal L(X^j_t\mid\F^0_t)$ for the conditional law from which $S^j_t$
is elicited, and from Assumptions~\ref{asm:lip_s} and~\ref{asm:lip_statistic}, we have that the coefficients are Lipschitz continuous with respect to the quadratic Wasserstein distance $\mathcal W_2$.

\smallskip
\noindent\emph{Two a priori estimates on $I_i$.} Let $(X,Y,Z,Z^0)$ and
$(X',Y',Z',Z^{0\prime})$ solve the sub-system on $I_i$ under the same Brownian
paths, with environments $S,S'$ (elicited from the laws $\mu,\mu'$) and boundary
data $(X_{t_i},Y_{t_{i+1}})$, $(X'_{t_i},Y'_{t_{i+1}})$.

\noindent\emph{Synchronous coupling.} Conditionally on $\F^0_t$, the conditional joint law of the pair 
$(X^k_t,X^\star_t)$ is a coupling of $(\mu^k_t,\mu^\star_t)$, whence
$\mathcal W_2(\mu^k_t,\mu^\star_t)^2\le\E\left[\norm{X^k_t - X^\star_t}^2\mid \F^0_t\right]$; taking
expectations, for every $k\ge0$,
\begin{equation}
  \E\big[\mathcal W_2(\mu^k_t,\mu^\star_t)^2\big]\;\le\;\E\left[\norm{X^k_t - X^\star_t}^2 \right] \le \mathcal{E}^k_i ,
  \qquad t\in I_i .
  \label{eq:synch}
\end{equation}

\smallskip
\noindent\emph{Forward step.} In \eqref{eq:apriori-fwd}, take the
unprimed $X_t$ solution to be the $k$-th forward iterate $X^k$, whose inputs are frozen
at level $k-1$ -- the previous environment $S^{k-1}$ and the previous backward
triple $(Y^{k-1},Z^{k-1},Z^{0,k-1})$ -- and the primed $X'_t$ solution to be $X^\star$.
The three right-hand terms of \eqref{eq:apriori-fwd} become:

\begin{enumerate}
\item $\E\big[\norm{X^k_{t_i} - X^\star_{t_i}}^2\big]=\delta^k_i$.
\vspace{5pt}
\item As $X^k$ is driven by $\Theta^{k-1}$ and $X^\star_t$ by $\Theta^\star_t$, 
\[\E\!\left[\int_{I_i}\!\norm{\Theta^{k-1}_t-\Theta^\star_t}^2\,{{\rm d} t}\right] = \B^{k-1}_i .\]
\vspace{5pt}
\item  Only $S^{k-1}$ vs.\ $S^\star$
differs, so by equations~\eqref{eq:drift_wass_lipschitz} and~\eqref{eq:synch}, we have 
\[
\begin{aligned}
  \Gamma\,\E\!\left[\int_{I_i}\!\big\|\mu(t,X_t,\Theta_t,S^{k-1}_t)
     -\mu(t,X_t,\Theta_t,S^\star_t)\big\|^2 {{\rm d} t}\right]
  &\;\le\; \Gamma L_m^2\!\int_{I_i}\!\E\big[\mathcal W_2(\mu^{k-1}_t,\mu^\star_t)^2\big]{{\rm d} t}\\
  &\;\le\; \Gamma L_m^2 h\,\mathcal E^{k-1}_i .
\end{aligned}
\]
\end{enumerate}
Collecting the three contributions,
\begin{equation}
  \mathcal E^k_i
  \;\le\; \Gamma\,\delta^k_i
        \;+\; \Gamma h\,\B^{k-1}_i
        \;+\; \Gamma L_m^2 h\,\mathcal E^{k-1}_i .
  \label{eq:forward_sde-sub}
\end{equation}

\smallskip
\noindent\emph{Backward step.} In \eqref{eq:apriori-bwd}, take the
unprimed solution $\Theta$ to be $\Theta^k$, which is driven by $(X^k,S^k)$ and has terminal datum $Y^k_{t_{i+1}}$, and the primed $\Theta'$ to be $\Theta^\star$; the left-hand side of \eqref{eq:apriori-bwd} is $\B^k_i$. The two right-hand terms are:
\begin{enumerate}
\item $\E\norm{Y^k_{t_{i+1}}-Y^\star_{t_{i+1}}}^2=\tau^k_i$.
\item The driver $f$
arguments differ only by $(X^k_t,S^k_t)$ vs.\ $(X^\star_t,S^\star_t)$; by
the Lipschitz bound in the $X_t$ and law slots, \eqref{eq:synch} with
$j=k$, and the $\sup$-bound on both resulting integrals,
\[
\begin{aligned}
  \E\!\int_{I_i}\!\big\|f(t,X^k_t,\Theta^k_t,S^k_t)-f(t,X^\star_t,\Theta^k_t,S^\star_t)\big\|^2 {{\rm d} t}
  &\;\le\; 2 \!\int_{I_i}\!\E\big[L^2\norm{X^k_t - X^\star_t}^2+ L_m^2\mathcal W_2(\mu^k_t,\mu^\star_t)^2\big]{{\rm d} t}\\
  &\;\le\; 2 (L^2 + L_m^2)h\,\mathcal E^k_i .
\end{aligned}
\]
\end{enumerate}
Collecting,
\begin{equation}
  \B^{k}_i
  \;\le\; \Gamma\,\tau^{k}_i
        \;+\; 2\Gamma (L^2 + L_m^2) h\,\mathcal E^{k}_i .
  \label{eq:backward_sde-bound}
\end{equation}

\smallskip
Substituting
\eqref{eq:backward_sde-bound} for $\B^{k-1}_i$ into
the backward term of \eqref{eq:forward_sde-sub},
\[
\begin{split}
  \mathcal E^k_i
  \;\le\; \Gamma\,\delta^k_i
        \;+\; \Gamma h\,\left(\Gamma\,\tau^{k-1}_i\;+\; 2\Gamma (L^2 + L_m^2) h\,\mathcal E^{k-1}_i \right)
        \;+\; \Gamma L_m^2 h\,\mathcal E^{k-1}_i\\
        = \Gamma\,\delta^k_i
        \;+\; \Gamma^2 h \tau^{k-1}_i\;+\; 
        \Gamma h \left(\,2\Gamma h L^2 + L_m^2( 1 +2\Gamma h)\,
        \right) \mathcal E^{k-1}_i,
\end{split}
\]
which is the stated inequality in the lemma, \eqref{eq:onesweep} with 
$\lambda(h)=\Gamma h \left(\,2\Gamma h (L^2 + L_m^2) + L_m^2\,\right) $. 
Finally, as $\lambda(h)\xrightarrow{h\downarrow0}0$, there exists $h_0$, s.t. $\lambda(h)<1$ whenever $h<h_0$; equivalently, for fixed $T$, it suffices to choose $N>\frac{T}{h_0}$.
\end{proof}

\subsection{Global Error Bound}\label{ssec:global_bound}

\begin{theorem}[Whole-interval bound with carried boundary errors]
\label{thm:global}
Under the standing idealization and Assumptions~\ref{asm:lip_s}-\ref{asm:lip_statistic}, with $h=T/N$ and
$\lambda(h)=\Gamma h \left(\,2\Gamma h (L^2 + L_m^2) + L_m^2\,\right) $ 
as in Lemma~\ref{lem:decay}, define the boundary-error budgets
\[
  D^j := \sum_{i=0}^{N-1}\delta^j_i
       = \sum_{i=0}^{N-1}\E\norm{X^j_{t_i} - X^\star_{t_i}}^2,
  \qquad
  \mathcal T^{j} := \sum_{i=0}^{N-1}\tau^{j}_i
       = \sum_{i=0}^{N-1}\E\norm{  Y^{j}_{t_{i+1}} - Y^{\star}_{t_{i+1}}  }^2 .
\]
Then for every $k\ge1$,
\begin{equation}
  \E\!\Big[\sup_{t\in[0,T]}\norm{X^k_t-X^\star_t}^2\Big]
  \;\le\;
  \lambda(h)^k\sum_{i=0}^{N-1}\mathcal E^0_i
  \;+\;
  \sum_{j=1}^{k}\lambda(h)^{\,k-j}
     \Big(\Gamma\,D^{j}+\Gamma^2 h\,\mathcal T^{\,j-1}\Big).
  \label{eq:wholebound}
\end{equation}
\end{theorem}

\begin{remark}    
The first term is the homogeneous decay of the initial-iteration error at the rate $\lambda(h)$. The second is the accumulation of the
boundary errors injected at each iteration -- the initial condition errors $D^j$ at the interior $t_i$ and the terminal condition errors $\mathcal T^{j-1}$
at $t_{i+1}$, each damped by the same geometric
factor $\lambda(h)$.
\end{remark}

\begin{proof}
At fixed $i$, \eqref{eq:onesweep} is an affine scalar recursion in $k$ with
multiplier $\lambda(h)$ and source $\Gamma\delta^k_i+\Gamma^2 h\,\tau^{k-1}_i$;
iterating it, we have that
\begin{equation}
\label{eqn:forward-error-iterate}
  \mathcal E^k_i
  \;\le\; \lambda(h)^k\,\mathcal E^0_i
   + \sum_{j=1}^{k}\lambda(h)^{\,k-j}\big(\Gamma\,\delta^j_i+\Gamma^2 h\,\tau^{j-1}_i\big). 
\end{equation}
As $[0,T]=\bigcup_{i=0}^{N-1}I_i$, one has pathwise 
\[
\sup_{t\in[0,T]}\|X_t^k - X_t^\star\|^2\le \max_{0\le i\le N-1}\sup_{t\in I_i} \| X_t^k - X_t^\star\|^2 \le \sum_{i=0}^{N-1}\sup_{t\in I_i}\| X_t^k - X_t^\star\|^2
\]
Taking expectations, bounding each term by $\mathcal E^k_i$ in \eqref{eqn:forward-error-iterate}, and summing over $i$ yields the stated result \eqref{eq:wholebound}.
\end{proof}

\begin{theorem}[Whole-interval bound for the backwards terms]
\label{thm:global-bwds}
Under the standing idealization and Assumptions~\ref{asm:lip_s}-\ref{asm:lip_statistic}, with $h=T/N$ and
$\lambda(h)=\Gamma h \left(\,2\Gamma h (L^2 + L_m^2) + L_m^2\,\right) $ 
as in Lemma~\ref{lem:decay}, $D^j$ and $\mathcal{T}^j$ as in Theorem~\ref{thm:global},
we have for all $k \geq 1$,
\begin{equation}
\begin{split}
  &\E\!\Big[\int_0^T\norm{\Theta^k_t-\Theta^\star_t}^2 {{\rm d} t}\Big]
  \;\le\; \Gamma\,\mathcal{T}^k\\
  &
        \;+\; 2\Gamma (L^2 + L_m^2) h\,
  \left(\lambda(h)^k\sum_{i=0}^{N-1}\mathcal E^0_i
  \;+\;
  \sum_{j=1}^{k}\lambda(h)^{\,k-j}
     \Big(\Gamma\,D^{j}+\Gamma^2 h\,\mathcal T^{\,j-1}\Big) \right).
  \label{eq:wholebound-backward}
\end{split}
\end{equation}

\end{theorem}

\begin{proof}
By inequalities~\eqref{eq:backward_sde-bound} and~\eqref{eqn:forward-error-iterate}, we have
\begin{equation}
  \B^{k}_i
  \;\le\; \Gamma\,\tau^{k}_i
        \;+\; 2\Gamma (L^2 + L_m^2) h\,
        \left( \lambda(h)^k\,\mathcal E^0_i
   + \sum_{j=1}^{k}\lambda(h)^{\,k-j}\big(\Gamma\,\delta^j_i+\Gamma^2 h\,\tau^{j-1}_i\big) \right).
\end{equation}
Summing over $i$ yields the result.
\end{proof}

\subsection{The inexact solver}
\label{sec:inexact}
We next drop the standing idealization. The networks $S_\theta,U_\theta,V_\theta,V^0_\theta$ no
longer return the exact conditional statistic and decoupling field: at every iteration
$k$ they use the samples of $\widehat X^k$ to return approximations $\widehat S^{\,k}$ and
$\widehat\Theta^{\,k}=(\widehat Y^{k},\widehat Z^{k},\widehat Z^{0,k})$ of the exact
statistic $S^{k}$ and the exact backward triple $\Theta^{k}$ -- the one produced by the
decoupling field of the sub-problem with the measure-flow argument frozen at $S^{k}$ --
each carrying a training and finite-capacity error. Here $X^k$ denotes the \emph{exact}
Picard iterate, driven by $\Theta^{k-1}$, and $\widehat X^k$ denotes the iterate driven by
the approximation $\widehat\Theta^{k-1}$. Since the networks are trained on samples of
$\widehat X^k$ rather than of $X^k$, we let the \emph{network error\nocorr} depend on the
discrepancy between the two:

\begin{assumption}\label{asm:network_error}    
There is a nondecreasing function $\enet:[0,\infty)\to[0,\infty)$ such that, for every
iteration $k\ge0$,
\begin{equation}
\begin{aligned}
  & \E\!\Big[\sup_{t\in [0,T]}\norm{\widehat S^{\,k}_t-S^{k}_t}^2\Big]\le\enet\big(\bar\varepsilon^{k}\big),  
  & & \E \left[\sup_{t\in [0,T]}\norm{\widehat Y^{\,k}_t-Y^{k}_t}^2\right] \le \enet\big(\bar\varepsilon^{k}\big),\\ 
  & \E\left[\int_0^T \norm{\widehat Z^{\,k}_s-Z^{k}_s}^2 \ds \right] \le \enet\big(\bar\varepsilon^{k}\big),
 & &\E \left[\int_0^T\norm{\widehat Z^{0,k}_s-Z^{0,k}_s}^2 \ds \right]\le\enet\big(\bar\varepsilon^{k}\big), \label{eq:netfloor}
\end{aligned}
\end{equation}
where
\begin{equation*}
  \bar\varepsilon^{k} \coloneqq \E\!\Big[\sup_{t\in [0,T]}\norm{\widehat X^{k}_t-X^{k}_t}^2\Big].
\end{equation*}
\end{assumption}
\noindent This assumption is motivated by universal approximation
theorems \cite{hornik1991approximation,pinkus1999approximation}: $\enet(0)$ is the training
and finite-capacity error, and the growth of $\enet$ accounts for the shift of the training
samples from $X^k$ to $\widehat X^k$. It is also motivated by functional dependence on $X$ for the inexact Picard solver error assumed in~\cite{chassagneux2019numerical}. A constant $\enet$ corresponds to a uniform network
error, and the idealization of Sections~\ref{ssec:local_bound}--\ref{ssec:global_bound} is
the case $\enet\equiv0$. 

Of particular interest is the case when $\enet$ satisfies
\begin{equation}\label{eq:affine-net-intro}
  \enet(x)\;\le\;\varepsilon_0+\ell\,x, \qquad x\ge0,
\end{equation}
When $x=0$, we are left with the network approximation error for the same input, represented by $\varepsilon_0$.
However, when $\varepsilon_0 = 0$ the network approximation error vanishes and the error is similar in form to Lipschitz continuity with respect to $X$. Therefore, this form can be interpreted as decomposing the network approximation error and the error due to the inputs.
We consider this case separately in Corollary~\ref{thm:affine_net}.

Define the forward error on $I_i$
\begin{equation*}
  \varepsilon^k_i \;\coloneqq\;
  \E\!\left[\sup_{t\in I_i}\big\|\widehat X^k_t-X^k_t\big\|^2\right].
\end{equation*}

\begin{lemma}\label{error_bound:inexact_forward_lemma}
 For every $0 \le i \le N-1$ and every $k\ge1$, we have
\begin{equation}
  \varepsilon^k_i \;\le\;  h\left(h + 2 + L_s^2 \right) \enet\big(\bar\varepsilon^{k-1}\big) \frac{\Gamma \left( \Gamma^{i+1} - 1 \right)}{\Gamma -  1} \le h\left(h + 2 + L_s^2 \right) \enet\big(\bar\varepsilon^{k-1}\big) \frac{\Gamma \left( \Gamma^{N} - 1 \right)}{\Gamma -  1}.
\end{equation}
Here $\Gamma > 1$ is assumed; for $\Gamma = 1$ the geometric factors
$\Gamma(\Gamma^{i+1}-1)/(\Gamma-1)$ and $\Gamma(\Gamma^{N}-1)/(\Gamma-1)$ are read as
their limits $i+1$ and $N$.
\end{lemma}

\begin{proof}
Using~\eqref{eq:apriori-fwd} with $X=\widehat X^k$, $X'=X^k$, we have
\begin{equation*}
  \Gamma\,\E\!\left[\big\|\widehat X^k_{t_i}-X^k_{t_i}\big\|^2\right]
  \;\le\; \Gamma\,\varepsilon^k_{i-1},
\end{equation*}
with the convention $\varepsilon^k_{-1}:=0$, which holds because
$\widehat X^k_0 = X^k_0 = X_0$ and which gives the base case $i=0$ of the induction below.
Note that the bounds in~\eqref{eq:netfloor} at iteration $k-1$ also hold on each $I_i$.
Therefore,
\begin{align*}
  & \E\!\left[\int_{I_i}\big\|\widehat\Theta^{k-1}_t-\Theta^{k-1}_t\big\|^2 {{\rm d} t}\right]
  \;\le\;  \E \left[h\,\sup_{t\in I_i}\norm{\widehat Y^{\,k-1}_t-Y^{k-1}_t}^2 \right] \\
  &+ \E \left[ \int_{I_i} \left(\norm{\widehat Z^{\,k-1}_t-Z^{k-1}_t}^2 + \norm{\widehat Z^{0,k-1}_t-Z^{0,k-1}_t}^2 \right) \dt\right] \;\le\; (h + 2)\,\enet\big(\bar\varepsilon^{k-1}\big). 
\end{align*}
Finally, using Assumption~\ref{asm:lip_s} and Equation~\eqref{eq:netfloor},
\begin{align*}
  \Gamma\,\E\!\left[\int_{I_i}
    \big\|\mu\big(t,\widehat X^k_t,\widehat\Theta^{k-1}_t,\widehat S^{k-1}_t\big)
        -\mu\big(t,\widehat X^k_t,\widehat\Theta^{k-1}_t,S^{k-1}_t\big)\big\|^2 {{\rm d} t}\right]
  &\le \Gamma\, L_s^2 \, \E\!\left[\int_{I_i} \big\|\widehat S^{k-1}_t-S^{k-1}_t\big\|^2 {{\rm d} t}\right]\\
  &\le \Gamma\, L_s^2 \, \E\!\left[h\sup_{t\in I_i}\big\|\widehat S^{k-1}_t-S^{k-1}_t\big\|^2\right]\\
  &\le \Gamma h L_s^2\,\enet\big(\bar\varepsilon^{k-1}\big) .
\end{align*}
Collecting the three bounds,
\begin{equation}\label{eq:rec}
  \varepsilon^k_i \;\le\;
  \Gamma\,\varepsilon^k_{i-1} + \underbrace{\Gamma h\,(h + 2)\,\enet\big(\bar\varepsilon^{k-1}\big) + \Gamma h\,L_s^2\, \enet\big(\bar\varepsilon^{k-1}\big)}_{=\Gamma h (h + 2 + L_s^2)\enet(\bar\varepsilon^{k-1})}.
\end{equation}
Therefore, by induction, it holds for all $i$, 
\begin{align*}
  \varepsilon^k_i \;\le\;
  \Gamma\,\varepsilon^k_{i-1} + \Gamma h (h + 2 + L_s^2)\enet\big(\bar\varepsilon^{k-1}\big) &\le \Gamma \left( h \left(h + 2 +  L_s^2 \right) \enet\big(\bar\varepsilon^{k-1}\big)  \frac{\Gamma \left( \Gamma^{i} - 1 \right)}{\Gamma -  1} \right ) \\
  &+ \Gamma h (h + 2 + L_s^2) \enet\big(\bar\varepsilon^{k-1}\big) \\
  &= h(h + 2 + L_s^2) \enet\big(\bar\varepsilon^{k-1}\big) \frac{\Gamma \left( \Gamma^{i+1} - 1 \right)}{\Gamma -  1}.
\end{align*}
By monotonicity of the expression in $i$ (since $\Gamma \geq 1$), we conclude the proof.
\end{proof}

\begin{theorem}[Whole-interval bound under an inexact solver]
\label{thm:global_inexact}
Under Assumptions~\ref{asm:lip_s}, \ref{asm:lip_statistic} and \ref{asm:network_error}, with $h=T/N$ and
$\lambda(h)<1$, set
\begin{equation*}
  C_N \coloneqq N h\left(h + 2 + L_s^2 \right) \frac{\Gamma \left( \Gamma^{N} - 1 \right)}{\Gamma -  1},
  \qquad
  \varphi_N(x) \coloneqq C_N\,\enet(x), \quad x\ge0,
\end{equation*}
and let $\varphi_N^{\circ k}$ denote the $k$-fold composition of $\varphi_N$, with
$\varphi_N^{\circ 0}$ the identity. Then the outer Picard iterates satisfy for every $k\ge1$
\begin{equation}\label{eq:global-error-bound-inexact}
\begin{aligned}
  &\E\!\Big[\sup_{t\in[0,T]}\norm{\widehat X^k_t-X^\star_t}^2\Big]
  \;\le\; 2\lambda(h)^k\sum_{i=0}^{N-1}\mathcal E^0_i\\
  & 
  \;+\;
  2\sum_{j=1}^{k}\lambda(h)^{\,k-j}
     \Big(\Gamma\,D^{j}+\Gamma^2 h\,\mathcal T^{\,j-1}\Big) + 
   2\,\varphi_N^{\circ k}\big(\bar\varepsilon^{0}\big).
\end{aligned}
\end{equation}
The first two terms are the idealized bound \eqref{eq:wholebound}; the third is the
propagated network error. For $\enet\equiv0$ we recover Theorem~\ref{thm:global} up to the
factor $2$ coming from~\eqref{eq:triangle-inexact}.
\end{theorem}

\begin{proof}
By the triangle inequality and $(a+b)^2 \le 2a^2 + 2b^2$, we have
\begin{equation}\label{eq:triangle-inexact}
    \sup_{t\in[0,T]}\norm{\widehat X^k_t-X^\star_t}^2 \leq 2\sup_{t\in[0,T]}\norm{\widehat X^k_t - X^k_t}^2 + 2\sup_{t\in[0,T]}\norm{X^k_t - X^\star_t}^2.
\end{equation}
For the first term we argue as in the proof of Theorem~\ref{thm:global}: the pathwise
bound $\sup_{t\in[0,T]}\|\cdot\|^2\le\sum_{i=0}^{N-1}\sup_{t\in I_i}\|\cdot\|^2$ and
Lemma~\ref{error_bound:inexact_forward_lemma} give, for every $j\ge1$,
\[
  \bar\varepsilon^{j}
  \;\le\; \sum_{i=0}^{N-1}\varepsilon^j_i
  \;\le\; C_N\,\enet\big(\bar\varepsilon^{j-1}\big)
  \;=\; \varphi_N\big(\bar\varepsilon^{j-1}\big).
\]
Since $\enet$, and hence $\varphi_N$, is nondecreasing, induction on $j$ gives
\begin{equation}\label{eq:forward-gap-iterate}
  \bar\varepsilon^{k} \;\le\; \varphi_N^{\circ k}\big(\bar\varepsilon^{0}\big).
\end{equation}
The second term is bounded by Theorem~\ref{thm:global}, and the result follows.
\end{proof}

\begin{remark}
    Note that as $h = T/N$, with $T$ fixed, $C_N$ grows as $T\,\Gamma^{\frac{T}{h}}$. Thus, there is a trade-off between decreasing $\lambda(h)$ and increasing $C_N$, which is attenuated by decreasing the network error $\enet$. Since $\enet$ depends on the forward discrepancy, the network error may also accumulate across iterations through $\varphi_N$; Corollary~\ref{thm:affine_net} quantifies this when $\enet$ satisfies~\eqref{eq:affine-net}.
\end{remark}

\begin{theorem}[Whole-interval bound under an inexact solver for the backwards terms]
\label{thm:global_inexact_bwds}
Under Assumptions~\ref{asm:lip_s}, \ref{asm:lip_statistic} and \ref{asm:network_error}, with $h=T/N$,
$\lambda(h)<1$ and $\varphi_N$ as in Theorem~\ref{thm:global_inexact}, the outer Picard iterates satisfy for every $k\ge1$
\begin{equation}
\begin{split}
  &\E\!\Big[\int_0^T\norm{{\widehat\Theta}^k_t-\Theta^\star_t}^2 {{\rm d} t}\Big]
  \;\le\;
  2\Gamma\,\mathcal{T}^k
        \\ &\;+\; 4\Gamma (L^2 + L_m^2) h\,
  \left(\lambda(h)^k\sum_{i=0}^{N-1}\mathcal E^0_i
  \;+\;
  \sum_{j=1}^{k}\lambda(h)^{\,k-j}
     \Big(\Gamma\,D^{j}+\Gamma^2 h\,\mathcal T^{\,j-1}\Big) \right) \\
     &+ 2(2 + T)\,\enet\Big(\varphi_N^{\circ k}\big(\bar\varepsilon^{0}\big)\Big).
  \label{eq:wholebound-backward-inexact}
\end{split}
\end{equation}
\end{theorem}

\begin{proof}
    From~\eqref{eq:netfloor}, we have that
\begin{equation}
     \E \left[\int_0^T\norm{\widehat Y^{\,k}_s-Y^{k}_s}^2 {\rm d} s\right] \leq T\E \left[\sup_{t\in [0,T]}\norm{\widehat Y^{\,k}_t-Y^{k}_t}^2\right] \leq T \enet\big(\bar\varepsilon^{k}\big).
\end{equation}
Gathering this with the inequalities in~\eqref{eq:netfloor} for $Z$ and $Z^0$, each of
which contributes $\enet(\bar\varepsilon^{k})$, and using~\eqref{eq:forward-gap-iterate}
together with the monotonicity of $\enet$, we get:
\begin{equation}
     \E \left[\int_0^T\norm{\widehat \Theta^{\,k}_s-\Theta^{k}_s}^2 \ds\right] \leq  (2+ T)\, \enet\big(\bar\varepsilon^{k}\big) \leq (2+ T)\, \enet\Big(\varphi_N^{\circ k}\big(\bar\varepsilon^{0}\big)\Big).
\end{equation}
Finally, by the triangle inequality, we have
\begin{equation}
    \int_0^T\norm{\widehat \Theta^k_s-\Theta^\star_s}^2 \ds \leq 2\int_0^T\norm{\widehat \Theta^k_s - \Theta^k_s}^2 \ds + 2\int_0^T\norm{\Theta^k_s - \Theta^\star_s}^2 \ds.
\end{equation}
From Theorem~\ref{thm:global-bwds}, the result follows.
\end{proof}

Both bounds in Theorems~\ref{thm:global_inexact}--\ref{thm:global_inexact_bwds} depend on the term $\varphi_N^{\circ k}\big(\bar\varepsilon^{0}\big)$. We conclude this subsection with Corollary~\ref{thm:affine_net}, which studies this term when the network error satisfies Inequality~\eqref{eq:affine-net}.

\begin{corollary}[Affine network error]
\label{thm:affine_net}
Under the assumptions of Theorem~\ref{thm:global_inexact}, suppose that there are
$\varepsilon_0,\ell\ge0$ such that
\begin{equation}\label{eq:affine-net}
  \enet(x)\;\le\;\varepsilon_0+\ell\,x, \qquad x\ge0,
\end{equation}
and set $r\coloneqq C_N\,\ell$. Then for every $k\ge1$,
\begin{equation}\label{eq:affine-iterate}
  \varphi_N^{\circ k}\big(\bar\varepsilon^{0}\big)
  \;\le\; r^k\,\bar\varepsilon^{0} + C_N\,\varepsilon_0\sum_{j=0}^{k-1} r^{j},
\end{equation}
In particular:
\begin{enumerate}
  \item[(i)] if $\varepsilon_0=0$, then for every $k\geq 1$ we have 
  $\varphi_N^{\circ k}\big(\bar\varepsilon^{0}\big) \le2\,r^k\,\bar\varepsilon^{0},$
   which vanishes when $\widehat X^0=X^0$;
  \item[(ii)] if $r<1$, then for every $k\ge1$
  \begin{equation*}
    \varphi_N^{\circ k}\big(\bar\varepsilon^{0}\big)
    \;\le\; r^k\,\bar\varepsilon^{0} + \frac{C_N\,\varepsilon_0}{1-r},
  \end{equation*}
  which is bounded uniformly in $k$;
  \item[(iii)] if $r=1$, then $\varphi_N^{\circ k}\big(\bar\varepsilon^{0}\big)\le\bar\varepsilon^{0}+k\,C_N\,\varepsilon_0$.
\end{enumerate}
\end{corollary}

\begin{proof}
Let $\psi_N(x)\coloneqq C_N(\varepsilon_0+\ell x)$, so that $\varphi_N\le\psi_N$
by~\eqref{eq:affine-net}, and $\psi_N$ is nondecreasing. Clearly,
$\varphi_N^{\circ k}\le\psi_N^{\circ k}$. Since $\psi_N(x)=r\,x+C_N\varepsilon_0$ is affine, induction on $k$ gives
\begin{equation*}
  \psi_N^{\circ k}(x) = r^k\,x + C_N\,\varepsilon_0\sum_{j=0}^{k-1} r^{j},
\end{equation*}
which is~\eqref{eq:affine-iterate}. 

Item (i) is~\eqref{eq:affine-iterate} with $\varepsilon_0=0$,
together with $\bar\varepsilon^{0}=0$ when $\widehat X^0=X^0$. Items (ii) and (iii) follow
from $\sum_{j=0}^{k-1}r^j\le(1-r)^{-1}$ for $r<1$ and $\sum_{j=0}^{k-1}r^j=k$ for $r=1$.
\end{proof}

\subsection{Quantile interaction}\label{ssec:quantile_bound}
In this subsection, we consider the case of the quantile interaction, which is not covered by the previous results.
In Lemma~\ref{lem:quantile-holder}, we give conditions for the quantiles to be H\"older continuous with respect to $\mathcal{W}_2$. In Lemma~\ref{lem:decay-holder}, we show that H\"older continuity suffices to provide local error bounds. The bounds for the whole interval for the exact and inexact solvers follow in a straightforward manner.

\begin{lemma}[H\"older continuity of the $\alpha$-quantile in $\mathcal W_p$]
\label{lem:quantile-holder}
Let $p \ge 1$, $\alpha \in (0,1)$ and $\mu, \nu \in \mathcal P_p(\mathbb R)$. Denote by $F_\mu$ the CDF of measure $\mu$ and $q_\alpha(\mu)$ its $\alpha$-quantile, with $q_\alpha(\mu) = F_\mu^{-1}(\alpha)$.

Suppose there exist constants $c > 0$ and
$\eta \in \big(0, \min\{\alpha, 1-\alpha\}\big)$ such that each of $\mu$ and $\nu$ admits an
interval $[a,b] \subset \mathbb R$, possibly different for the two measures, satisfying
\begin{enumerate}
  \item[(i)] $F_\mu(a) < \alpha - \eta$ and $F_\mu(b) > \alpha + \eta$, and likewise for $F_\nu$;
  \item[(ii)] the restriction of the measure to $[a,b]$ is absolutely continuous, with
        density $f$ satisfying $f \ge c$ a.e.\ on $[a,b]$.
\end{enumerate}
Then
\begin{equation}
  \big|q_\alpha(\mu) - q_\alpha(\nu)\big|
  \;\le\;
  \max\left\{
    \left( \frac{(p+1)\, \mathcal W_p(\mu,\nu)^p}{c} \right)^{\frac{1}{p+1}},
    \;
    \left( \frac{(p+1)\, \mathcal W_p(\mu,\nu)^p}{2\eta} \right)^{\frac{1}{p}}
  \right\}.
  \label{eq:quantile-holder-global}
\end{equation}
If moreover
\begin{equation}
  \mathcal W_p(\mu,\nu)^p \;\le\; \frac{2^{\,p+1}\, \eta^{\,p+1}}{(p+1)\, c^{\,p}},
  \label{eq:quantile-holder-prox}
\end{equation}
then
\begin{equation}
  \big|q_\alpha(\mu) - q_\alpha(\nu)\big|
  \;\le\;
  \left( \frac{p+1}{c} \right)^{\frac{1}{p+1}}
  \mathcal W_p(\mu,\nu)^{\frac{p}{p+1}} .
  \label{eq:quantile-holder-local}
\end{equation}
In particular, for $p = 2$ and $\mathcal W_2(\mu,\nu)$ small enough,
\begin{equation}
  \big|q_\alpha(\mu) - q_\alpha(\nu)\big|
  \;\le\; \left( \frac{3}{c} \right)^{\frac13} \mathcal W_2(\mu,\nu)^{\frac23}.
\end{equation}
\end{lemma}

\begin{proof}
Write $I \coloneqq [\alpha - \eta, \alpha + \eta]$, and let $F$ be the distribution
function of either measure, with $[a,b]$ and $f$ as in (i)--(ii).
By (ii), $F$ is continuous on $[a,b]$ and, for $a \le x \le y \le b$,
\begin{equation*}
  F(y) - F(x) = \int_x^y f(z)\, {\rm d} z \;\ge\; c\,(y-x).
\end{equation*}
This implies that
\begin{equation}
  \big|F^{-1}(v) - F^{-1}(u)\big| \;\le\; \frac1c\,|v-u|,
  \qquad u,v \in I,
  \label{eq:quantile-lip}
\end{equation}
for both $\mu$ and $\nu$.

On the real line the monotone rearrangement is an optimal coupling for every $p \ge 1$, so
\begin{equation}
  \mathcal W_p(\mu,\nu)^p
  = \int_0^1 \big| F_\mu^{-1}(u) - F_\nu^{-1}(u) \big|^p \, \du ;
  \label{eq:w1d}
\end{equation}
see \cite{villani2003topics,santambrogio2015optimal}. Set 
\begin{align*}
&g \coloneqq F_\mu^{-1} - F_\nu^{-1},\\
&\delta \coloneqq |g(\alpha)| = |q_\alpha(\mu) - q_\alpha(\nu)| .
\end{align*}
Assume $\delta > 0$. For $u \in I$, the triangle inequality together
with~\eqref{eq:quantile-lip} applied to both measures gives
\begin{equation*}
  |g(u)|
  \;\ge\; |g(\alpha)|
        - \big|F_\mu^{-1}(u) - F_\mu^{-1}(\alpha)\big|
        - \big|F_\nu^{-1}(u) - F_\nu^{-1}(\alpha)\big|
  \;\ge\; \delta - \frac{2}{c}\,|u - \alpha| .
\end{equation*}

Let $r \in (0, \eta]$ with $r \le \frac{c\delta}{2}$. Restricting~\eqref{eq:w1d} to
$[\alpha - r, \alpha + r]$ we have,
\begin{equation}
  \mathcal W_p(\mu,\nu)^p
  \;\ge\; \int_{\alpha-r}^{\alpha+r} |g(u)|^p\, \du
  \;\ge\; 2\int_0^r \left( \delta - \frac{2t}{c} \right)^{p} {{\rm d} t} .
  \label{eq:integrated}
\end{equation}

\emph{Case 1: $\frac{c\delta}{2} \le \eta$.} Choosing $r = \frac{c\delta}{2}$ in~\eqref{eq:integrated} and
substituting $s = \delta - \frac{2t}{c}$,
\begin{equation*}
  \mathcal W_p(\mu,\nu)^p \;\ge\; 2 \cdot \frac{c}{2} \int_0^\delta s^p\, ds
  = \frac{c\,\delta^{\,p+1}}{p+1},
\end{equation*}
so that 
\begin{equation*}
    \delta \le \big((p+1)\mathcal W_p(\mu,\nu)^p / c\big)^{1/(p+1)}.
\end{equation*}

\emph{Case 2: $\frac{c\delta}{2} > \eta$.} Choose $r = \eta$. Since $\frac{2\eta}{c\delta} < 1$, for
$t \in [0,\eta]$ we have
\begin{equation*}
  \delta - \frac{2t}{c}
  = \delta\left( 1 - \frac{2\eta}{c\delta}\cdot\frac{t}{\eta} \right)
  \;\ge\; \delta\left( 1 - \frac{t}{\eta} \right),
\end{equation*}
and therefore
\begin{equation*}
  \mathcal W_p(\mu,\nu)^p
  \;\ge\; 2\delta^p \int_0^\eta \left( 1 - \frac{t}{\eta} \right)^p {{\rm d} t}
  = \frac{2\eta\,\delta^p}{p+1},
\end{equation*}
so that 
\begin{equation*}
    \delta \le \big((p+1)\mathcal W_p(\mu,\nu)^p / (2\eta)\big)^{1/p}.
\end{equation*}

Together the two cases give~\eqref{eq:quantile-holder-global}. Finally, assume~\eqref{eq:quantile-holder-prox} and suppose Case 2 occurs. Then
\begin{equation*}
  \delta \;\le\; \left( \frac{(p+1)\,\mathcal W_p(\mu,\nu)^p}{2\eta} \right)^{\frac1p}
  \;\le\; \frac{2\eta}{c},
\end{equation*}
contradicting $\frac{c\delta}{2} > \eta$. Hence only Case 1 occurs, which
is~\eqref{eq:quantile-holder-local}.
\end{proof}

\begin{lemma}[Local error bound up to boundary mismatches under H\"older continuity]
\label{lem:decay-holder}
Under the standing idealization and Assumptions~\ref{asm:lip} and~\ref{asm:lip_s},
suppose the statistic satisfies, with exponent $2/3$ and some $L_\mathcal{S} > 0$,
\begin{equation}\label{eq:holder-wrt-wass-statistic}
  \norm{\mathcal S(\mu) - \mathcal S(\mu')} \le L_\mathcal{S}\, \mathcal W_2(\mu,\mu')^{2/3}
\end{equation}
for all conditional laws $\mu,\mu'$ of the iterates and of the solution. 
Let $\Gamma = \Gamma(L) \ge 1$ be as in Lemma~\ref{lem:decay}, and set
\begin{equation*}
  a(h) \coloneqq 2\Gamma^2 h^2 L^2, \qquad
  b(h) \coloneqq \Gamma h\, L_s^2 L_\mathcal{S}^2 (1 + 2\Gamma h),
\end{equation*}
\begin{equation*}
  \rho(h) \coloneqq \frac{1 + a(h)}{2}, \qquad
  \kappa(h) \coloneqq \frac{16}{27}\, \frac{b(h)^3}{(1 - a(h))^2}.
\end{equation*}
If $a(h) < 1$, then for every $i \in \{0,\dots,N-1\}$ and every $k \ge 1$,
\begin{equation}
  \mathcal E^k_i \le \rho(h)\, \mathcal E^{k-1}_i + \Gamma\, \delta^k_i
  + \Gamma^2 h\, \tau^{k-1}_i + \kappa(h).
  \label{eq:onesweep-holder}
\end{equation}
In particular $\rho(h) < 1$ and $\kappa(h)\to 0$ as $h\downarrow0$.
\end{lemma}

\begin{remark}
    By
Lemma~\ref{lem:quantile-holder}, an $\alpha$-quantile statistic satisfies~\eqref{eq:holder-wrt-wass-statistic} as long as the laws involved stay in the regime of the
smallness condition~\eqref{eq:quantile-holder-prox}; we assume throughout that the
iterates remain in that regime, since otherwise only the weaker global
bound~\eqref{eq:quantile-holder-global} is available.
\end{remark}

\begin{proof}
The proof follows that of Lemma~\ref{lem:decay}, with the Wasserstein terms
treated as follows. By the H\"older bound, Jensen's inequality for the concave map
$x \mapsto x^{2/3}$, and~\eqref{eq:synch},
\begin{equation}
  \E\left[\norm{\mathcal S(\mu^j_t) - \mathcal S(\mu^\star_t)}^2\right]
  \le L_\mathcal{S}^2\, \E\big[\mathcal W_2(\mu^j_t, \mu^\star_t)^{4/3}\big]
  \le L_\mathcal{S}^2 \big(\E\big[\mathcal W_2(\mu^j_t, \mu^\star_t)^{2}\big]\big)^{2/3}
  \le L_\mathcal{S}^2\, (\mathcal E^j_i)^{2/3}
\end{equation}
for $j \in \{k-1, k\}$. Repeating the forward and backward steps with this bound gives
\begin{equation}
  \mathcal E^k_i \le \Gamma\, \delta^k_i + \Gamma^2 h\, \tau^{k-1}_i
  + a(h)\, \mathcal E^{k-1}_i + b(h)\, (\mathcal E^{k-1}_i)^{2/3}.
\end{equation}
By Young's inequality with conjugate exponents $3/2$ and $3$, for every $\varepsilon > 0$ and $x \ge 0$,
\begin{equation}
  b\, x^{2/3} \le \frac{2}{3}\, \varepsilon\, x + \frac{1}{3}\, b^3 \varepsilon^{-2}.
\end{equation}
Taking $\varepsilon = \tfrac34 (1 - a(h))$ yields~\eqref{eq:onesweep-holder}.
\end{proof}

\section{Numerical experiments}\label{sec:experiments}
In this section, we present numerical experiments to demonstrate the effectiveness of our methodology. The first experiment solves the systemic risk banking model introduced in \cite{carmona2013mean}. This model can be solved analytically, hence it is useful as a benchmark against our numerical methodology. 
Building on this foundation, our second experiment modifies the systemic risk model by replacing interaction through the \textit{mean} of the agents for an interaction through a \textit{quantile} of the population. 
Finally, our third experiment solves a non-stationary Aiyagari–Bewley–Huggett model of income and wealth distribution \cite{achdou2022income}, where the mean field interaction is mediated through the interest rate and the agents are subjected to a common noise.

The processes we approximate depend on the
driving randomness in two distinct ways, and we choose the network
architecture accordingly. The conditional statistic $S_t$ is adapted to the
common noise alone, so we represent it through a recurrent network $S_\theta$
acting on the common-noise path $(W^0_s)_{s \le t}$. The decoupling field
$U_\theta$ and the gradient networks $v_\theta$ and $v^0_\theta$ instead depend
on both the current state and the history of the common noise through their
arguments $(t, X_t, (W^0_s)_{s \le t})$; for these we use a hybrid architecture
with two parallel branches.

The Markovian inputs $(t, X_t)$ are mapped by a
feedforward layer, while the common-noise path is encoded by a single GRU
layer; the two representations are then concatenated along the feature
dimension and fused by a further feedforward layer that returns them to a
common width. The fused representation is subsequently refined by four residual
feedforward blocks---each a linear map with a SiLU activation and a skip
connection---after which a final linear map yields the output. All hidden
layers use $18$ units and the SiLU activation. 

Separating the inputs in this
way lets each network exploit the Markovian dependence on $(t, X_t)$ directly
while still capturing the genuinely path-dependent effect of the common noise,
and confines the recurrent component to the common-noise path, where that
path-dependence actually resides. These networks are small compared with many
architectures one might consider; nonetheless, for the experiments below they
prove expressive enough to recover the solutions accurately.

\begin{figure}[t]
  \centering
  \resizebox{\textwidth}{!}{%
\definecolor{cIO}{HTML}{4B5563}
\definecolor{cMark}{HTML}{2563EB}
\definecolor{cGRU}{HTML}{0D9488}
\definecolor{cOp}{HTML}{D97706}
\definecolor{cTrunk}{HTML}{7C3AED}
\begin{tikzpicture}[
  x=1mm, y=1mm,
  font=\sffamily,
  >={Stealth[round]},
  base/.style   = {rounded corners=2.5pt, draw, line width=1pt,
                   minimum height=13mm, minimum width=24mm,
                   align=center, inner sep=3pt, fill=white,
                   font=\sffamily\Large},
  io/.style     = {base, draw=cIO,    text=cIO,    font=\sffamily\Large\bfseries},
  markov/.style = {base, draw=cMark,  fill=cMark!8},
  gru/.style    = {base, draw=cGRU,   fill=cGRU!10},
  trunk/.style  = {base, draw=cTrunk, fill=cTrunk!8},
  op/.style     = {circle, draw=cOp, line width=1.2pt, fill=cOp!12,
                   inner sep=0pt, minimum size=10mm, font=\LARGE},
  act/.style    = {font=\small\sffamily\itshape, text=black!55},
  dim/.style    = {font=\small\sffamily, text=black!50},
  glab/.style   = {font=\small\sffamily\bfseries},
  flow/.style   = {->, line width=1pt, draw=black!65},
  skip/.style   = {->, line width=1pt, draw=cTrunk!70, dashed},
]
\node[io]     (inp)     at (0,0)    {Input $x$};
\node[dim, below=0.5mm of inp] {$(B,T,d_M{+}d_P)$};
\node[markov] (linm)    at (34,16)  {Linear\\$d_M \to h$};
\node[dim, above=0.5mm of linm] {Markov slice $x[:d_M]$};
\node[gru]    (gru)     at (34,-16) {GRU\\$d_P \to g$};
\node[dim, below=0.5mm of gru] {Path slice $x[d_M:]$ \; ($L_g$ layers)};
\node[op]     (cat)     at (64,0)   {$\oplus$};
\node[trunk]  (bridge)  at (92,0)   {Bridge\\$h{+}g \to h$};
\node[trunk]  (res)     at (126,0)  {Residual\\$h \to h$};
\node[dim, below=0.5mm of res] {$\times\,4$};
\node[op]     (add)     at (148,0)  {$+$};
\node[trunk]  (out)     at (176,0)  {Output\\$h \to d_{\text{out}}$};
\node[io]     (rout)    at (206,0)  {Output};
\node[dim, below=0.5mm of rout] {$(B,T,d_{\text{out}})$};
\draw[flow] (inp.east) -- ++(7,0) coordinate (sp) |- (linm.west);
\draw[flow] (sp) |- (gru.west);
\draw[flow] (linm.east) -| node[act, pos=0.18, above] {SiLU} ($(cat.north)+(0,6)$) -- (cat.north);
\draw[flow] (gru.east)  -| node[act, pos=0.18, below] {SiLU} ($(cat.south)-(0,6)$) -- (cat.south);
\node[dim, below right=0mm and 0.5mm of linm.east, yshift=-1mm] {$(B,T,h)$};
\node[dim, above right=0mm and 0.5mm of gru.east,  yshift=1mm]  {$(B,T,g)$};
\draw[flow] (cat.east)    -- (bridge.west);
\draw[flow] (bridge.east) -- node[act, above] {SiLU} (res.west);
\draw[flow] (res.east)    -- (add.west);
\draw[flow] (add.east)    -- node[act, above] {SiLU} (out.west);
\draw[flow] (out.east)    -- (rout.west);
\draw[skip] ($(bridge.east)+(5,0)$) -- ++(0,12) -| (add.north);
\begin{scope}[on background layer]
  \node[draw=cMark!35, line width=0.7pt, rounded corners=4pt, dash pattern=on 2pt off 2pt,
        fit=(linm), inner sep=5mm,
        label={[cMark, glab]above:Markov stream}] {};
  \node[draw=cGRU!40, line width=0.7pt, rounded corners=4pt, dash pattern=on 2pt off 2pt,
        fit=(gru), inner sep=5mm,
        label={[cGRU, glab]below:Path-dependent stream}] {};
  \node[draw=cTrunk!30, line width=0.7pt, rounded corners=4pt, dash pattern=on 2pt off 2pt,
        fit=(bridge)(res)(add)(out), inner sep=4mm,
        label={[cTrunk, glab]above:Shared trunk}] {};
\end{scope}
\end{tikzpicture}
\space
      }
  \caption{Architecture of the hybrid approximator.}
\end{figure}

Each Picard iteration consists of $E=2,000$ backpropagation iterations for the $Y, Z, Z^0$ and $S$ neural nets, using batches of $I=8,192$ paths. We use the AdamW optimizer with learning rate 0.0005 and a decay each 5 steps of 0.9997. Soft updates of neural networks in  \eqref{eq:soft_updates} are performed with $\delta =0.5$, following the fictitious play methodology described in \cite{lauriere2021numerical}. Moreover, we discretize the time interval in $N=101$ timesteps, sample $M=50,000$ paths to train the neural networks, and we perform $K = 20$ (outer) Picard iterations. Each (outer) Picard iteration takes on average 8 minutes and 12 seconds to execute on a T4 machine in \texttt{Google Colab}. 

\subsection{Systemic risk banking model}

\subsubsection{Problem formulation}
    As the model presented in \cite{carmona2013mean} is linear-quadratic, it admits a explicit solution and hence provides  useful testing ground.
    In the model, each bank controls its rate of borrowing/lending to a central bank.
    The state $X_t$ represents the log-monetary reserve of a representative bank, which reverts to the mean log-monetary reserve $S_t$.
    The Nash equilibrium is characterized by the solution to the MV-FBSDE system
    \begin{align}\label{systemic_risk:model}
    \left\{
        \begin{aligned}
            \dX_t &= \left[ (a + q) (S_t - X_t) - Y_t \right]\, \dt + \sigma\, \dB_t,  & X_0 = \xi,
            \\[0.25em]
            \dY_t &= \left[ (a + q) Y_t + (\epsilon - q^2) (S_t - X_t) \right]\, \dt + Z_t\, \dW_t + Z_t^0\, \dW_t^0, 
            & Y_T = c\,(X_T - S_T),
        \end{aligned}
    \right.
    \end{align} 
    where $B_t = \rho\,W_t^0 + \sqrt{1-\rho^2}\,W_t$ and $S$ is the conditional mean:
\begin{align}\label{mean_conditional_on_common_noise}
    S_t = \E\left[ X_t \mid \F_t^0\,\right].
\end{align}
    The analytic solution for $Y_t$ is
    \begin{equation}
        Y_t = - \eta(t)\, (S_t - X_t),
    \end{equation}
    where $\eta$ is the solution to the Ricatti ODE
    \begin{equation}
        \dot \eta(t) = 2(a+ q) \eta(t) + \eta^2(t) - (\epsilon - q^2),
    \end{equation}
    which is given by
    \begin{equation}\label{systemic_risk:eta}
        \eta(t) = \frac{
             -(\epsilon - q^2) \left( e^{(\delta^+ - \delta^-) (T - t)} - 2 \right) 
              - c \left( \delta^+ e^{(\delta^+ - \delta^-) (T - t)} - \delta^- \right)
              }{
                \left( \delta^- e^{(\delta^+ - \delta^-) (T - t)} - \delta^+ \right)
                - c \left( e^{(\delta^+ - \delta^-) (T - t)} - 1 \right)
              },
    \end{equation}
    with
    \begin{equation}
        \delta^{\pm} = - (a + q) \pm \sqrt{ (a + q)^2 + (\epsilon - q^2) }.
    \end{equation}

    Hence, the dynamics for $X_t$ may be written
    \begin{align}\label{common_noise:state_dynamics}
        dX_t &= \left[ (a + q + \eta(t)) (S_t - X_t) \right]\, dt + \sigma\, \dB_t, & X_0 = \xi.
    \end{align}
    Applying conditional expectations  on the common noise on \eqref{common_noise:state_dynamics}, 
    we conclude that $S_t =\E[\xi] + \rho \,\sigma\, W^0_t$.
    Using the integrating factor $\Theta(t) = \int_0^t \theta(s) ds$, where $\theta(t) = a + q + \eta(t)$, the complete analytical solution for the MV-FBSDE is given by
\begin{equation}
    \begin{cases}
        \displaystyle S_t = \E[\xi] + \rho\, \sigma \,W^0_t,
        \\[0.25em]
       \displaystyle  X_t = \xi e^{-\Theta(t)} + \int_0^t \theta(u) S_u e^{-({\Theta(t) - \Theta(u)})}\, \du + \sigma \int_0^t e^{-({\Theta(t) - \Theta(u)})} \,\dB_u,
       \\[0.25em]
       \displaystyle  Y_t = - \eta(t) (S_t - X_t),
       \\[0.25em]
       \displaystyle Z_t = \sigma \,\eta(t),
       \\[0.25em]
       \displaystyle Z_t^0 = 0.
    \end{cases}
\end{equation}

\subsubsection{Numerical simulation and parameters}
For numerical experiments, we choose parameters values that appear in \cite{carmona2013mean}: $a = q = c = \sigma = 1$, $\epsilon = 10$, and $\rho=0.3$.  The initial condition $\xi$ is sampled from the normal distribution with mean $0$ and variance $4$. 

\begin{figure}[h!]

\includegraphics[width=1\textwidth]{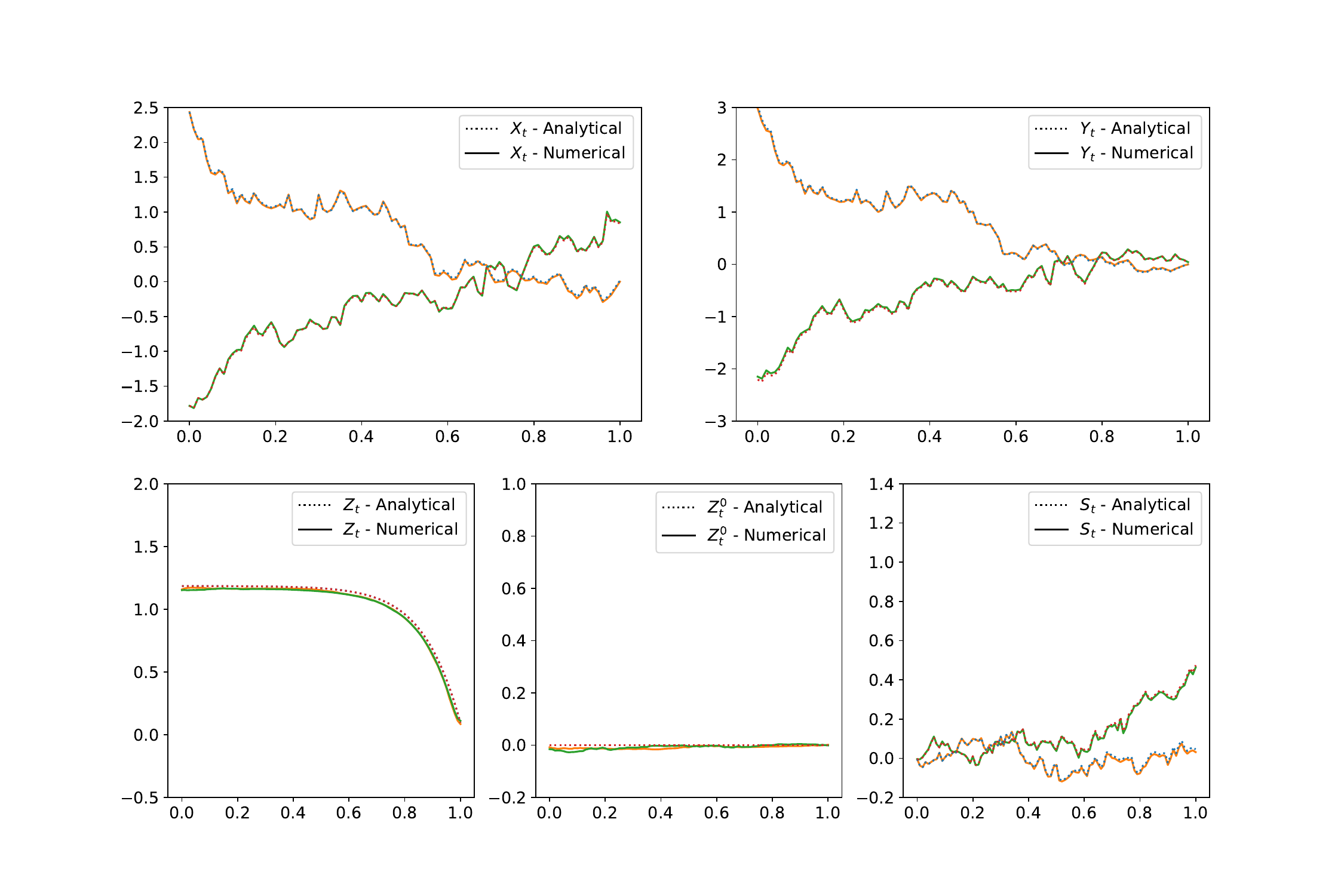}
\caption[Sample Paths - Systemic Risk With Common Noise]{
Comparison between analytic solution and approximated solution on two sample paths for the systemic risk with common noise experiment. Solid lines are the numerical solutions, and dotted lines represent the analytical solution. Each color pair (orange and blue, green and red) represents a particular path of each variable of the system.
}
\label{systemicRiskCommonNoise:SamplePaths}
\end{figure}

\begin{figure}[h!]

\includegraphics[width=1\textwidth]{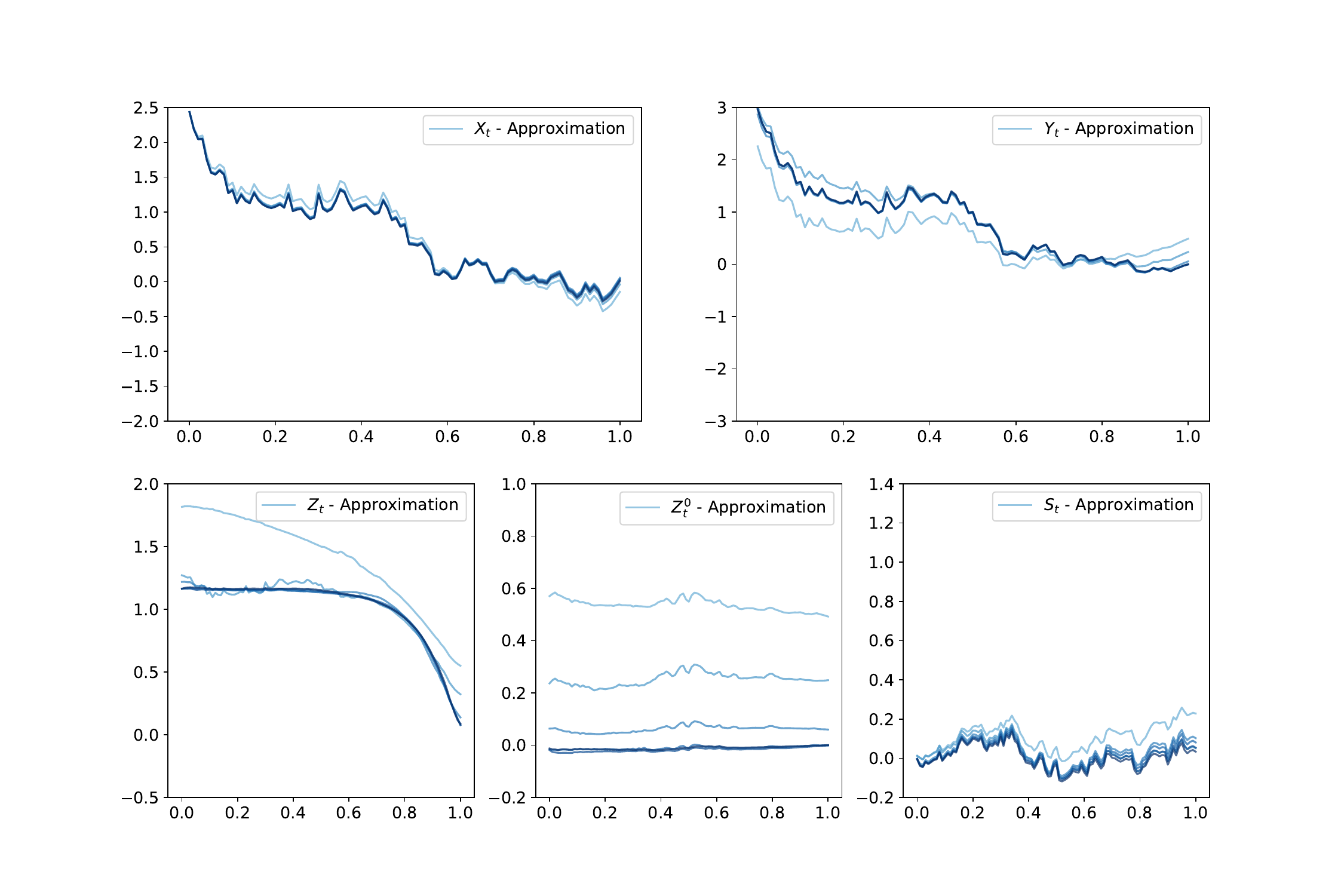}
\caption[Convergence of Iterations]{
The same sample path at iterations 1,2,4,8,16 and 32. As iteration increases, the color becomes darker.
}
\label{systemicRiskCommonNoise:Iterations}
\end{figure}

Figure \ref{systemicRiskCommonNoise:SamplePaths}
shows a comparison between approximated and analytical solutions for two realizations, while, in Figure 
\ref{systemicRiskCommonNoise:Iterations}, we show a sample path as the outer Picard iterations evolve. 

\subsubsection{Increasing the number of dimensions}
In table \ref{tab:mee}, we make a quantitative comparison of the mean Euclidean error\footnote{The Mean Euclidean Error (MEE) is defined as the average Euclidean norm over time and paths $\text{MEE}(x,y) = \frac{1}{|M||\mathcal{T}|}\sum_{0\leq i \leq M, t \in \mathcal{T}} \sqrt{(x^{(i)}_t - y^{(i)}_t)^2}$.} (MEE) per iteration as the number of dimensions $d$ range between $1$, $5$ and $10$. In this experiment, each dimension is an independent copy of the systemic risk banking model \eqref{systemic_risk:model}, with the same set of parameters. 

In this experiment we modify the size of the neural networks, the training setup, and include a resampling procedure. The number of hidden layers for the path-dependent GRU is set to be equal to $d+2$, and the feed-forward shared trunk is composed of $3$ layers of $48$ units.
The number of time steps is increased to $|\tau| = 301$. At each Picard Iteration, we perform $100$ SGD steps of the neural networks for $Y_t, Z_t, Z^0_t, S_t$. However, we increase the number of Picard Iterations to $250$ - for a total of $25,000$ SGD steps. Each SGD step is computed in a batch of $I =8,192$ paths, sampled uniformly with replacement from a set of $M=50,000$. At every $20$-th Picard Iteration, we resample the $50,000$ paths by means of Algorithm~\ref{methodology:sampling_after_training}. 

From the table, we can conclude that the algorithm succeeds in approximating the Nash equilibrium of \eqref{systemic_risk:model} for all 3 settings, even though there is an expected decrease in performance as the number of dimensions grow.
In comparison with \cite{Hu2025}, the MEE are of a similar order of magnitude.

\begin{table}[H]
\centering
\small

\begin{tabular}{
cc
S[table-format=2.2]@{ (}l@{)\hspace{1.2em}}
S[table-format=2.2]@{ (}l@{)\hspace{1.2em}}
S[table-format=2.2]@{ (}l@{)\hspace{1.2em}}
S[table-format=2.2]@{ (}l@{)\hspace{1.2em}}
S[table-format=2.2]@{ (}l@{)}
}
\toprule
$d$ & $K$
& \multicolumn{2}{c}{$X$}
& \multicolumn{2}{c}{$Y$}
& \multicolumn{2}{c}{$Z$}
& \multicolumn{2}{c}{$Z^0$}
& \multicolumn{2}{c}{$S$} \\

\midrule
\multirow{4}{*}{$d=1$}
& $k=5$
& 3.54  & 1.99
& 8.22  & 5.97
& 16.80 & 11.2
& 19.70 & 3.54
& 3.82  & 2.03 \\

& $k=50$
& 0.54 & 0.46
& 1.08 & 1.00
& 2.08 & 2.18
& 3.84 & 1.84
& 0.45 & 0.56 \\

& $k=100$
& 0.59 & 0.44
& 1.53 & 0.94
& 1.71 & 1.30
& 1.76 & 1.11
& 0.36 & 0.45 \\

& $k=250$
& 0.41 & 0.34
& 0.81 & 0.81
& 1.59 & 1.13
& 0.60 & 0.60
& 0.47 & 0.44 \\

\midrule
\multirow{4}{*}{$d=5$}
& $k=5$
& 11.10 & 6.94
& 31.50 & 17.2
& 45.20 & 25.0
& 45.40 & 7.45
& 12.60 & 8.18 \\

& $k=50$
& 2.29 & 1.35
& 4.95 & 2.92
& 3.67 & 3.09
& 5.17 & 2.21
& 3.01 & 2.08 \\

& $k=100$
& 2.29 & 1.23
& 4.73 & 2.75
& 3.46 & 2.42
& 4.16 & 1.63
& 2.82 & 1.83 \\

& $k=250$
& 1.79 & 0.91
& 3.65 & 2.07
& 2.38 & 1.74
& 3.27 & 1.22
& 2.18 & 1.37 \\

\midrule
\multirow{4}{*}{$d=10$}
& $k=5$
& 21.30 & 11.7
& 60.10 & 31.3
& 58.20 & 35.4
& 69.30 & 13.8
& 24.00 & 13.0 \\

& $k=50$
& 4.20 & 2.41
& 10.50 & 5.08
& 7.50 & 6.38
& 8.02 & 2.36
& 5.87 & 3.53 \\

& $k=100$
& 3.75 & 2.00
& 8.74 & 4.32
& 6.16 & 4.51
& 6.58 & 2.01
& 4.99 & 2.85 \\

& $k=250$
& 3.33 & 1.67
& 6.97 & 3.55
& 4.38 & 2.75
& 4.50 & 1.38
& 4.43 & 2.35 \\

\bottomrule
\end{tabular}

\caption{MEE results ($\times 10^{-2}$) for different values of $K$ and dimensions. In parentheses, the standard deviation of the Euclidean norm of errors by path and time.}
\label{tab:mee}
\end{table}

\subsection{Interaction through quantile}

\begin{figure}[h!]
\includegraphics[width=1\textwidth]{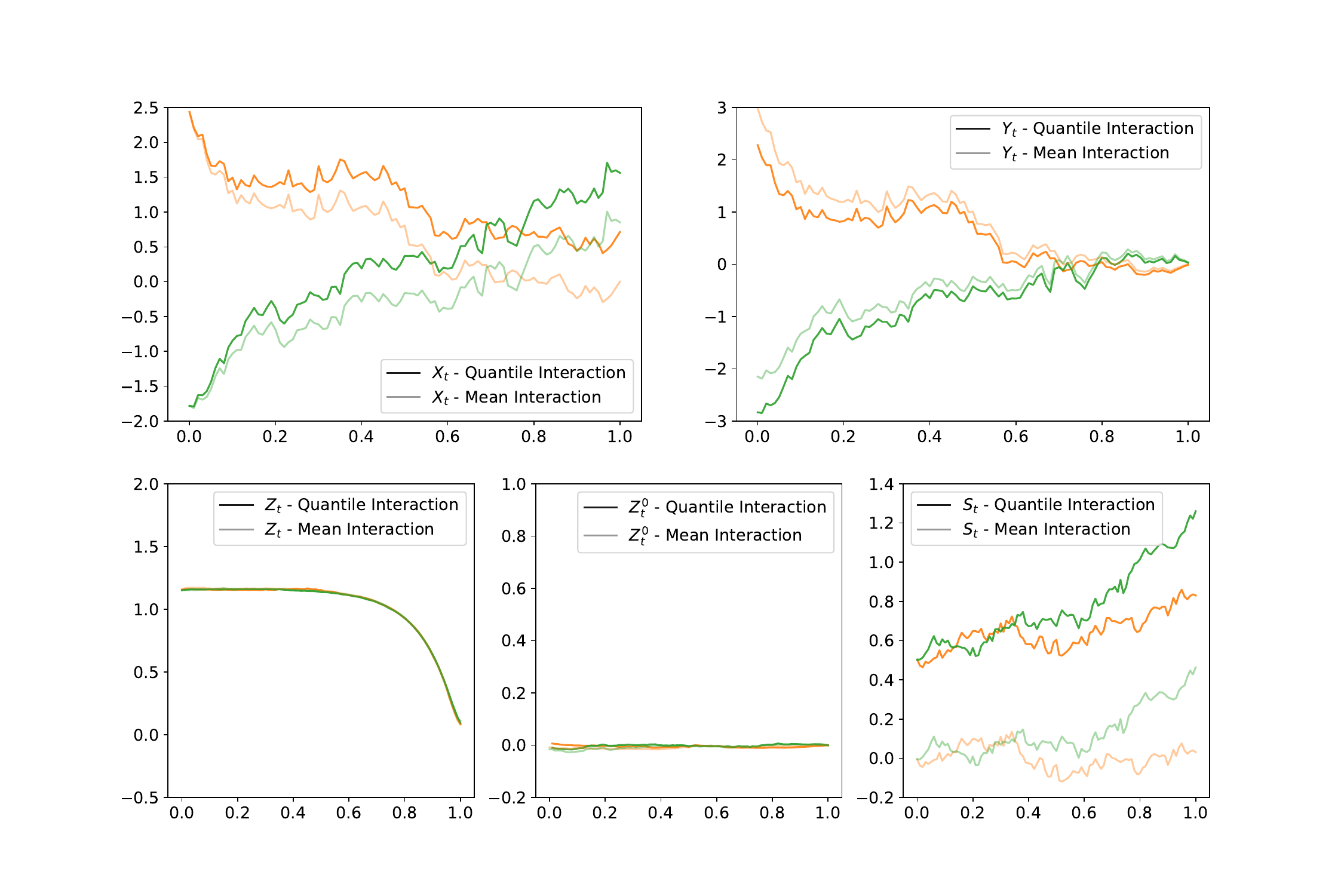}
\caption[Interaction Through Quantile]{Modified systemic risk model where interaction is mediated through the $60\%$ quantile. Respective paths (subjected to the same Brownian motion) of the original, mean interaction systemic risk model are shown in lighter shades. Note that the $X_t$ variable is greater in the quantile interaction model.}
\label{systemicRiskMedian:MeasureFlow}
\end{figure}

To test our algorithm in more challenging example, we apply the algorithm to a modified version of system \eqref{systemic_risk:model} where the interaction term $S_t$ is the $\alpha$-quantile, with $\alpha=60\%$, see Equation \eqref{eq:alpha_quantile}. We use the same values for all other parameters. Figure \ref{systemicRiskMedian:MeasureFlow} shows two sample paths of the variables at Nash equilibrium, along with the elicited quantile $\alpha=60\%$. The sample paths are subjected to the same Brownian motions $W_t$ and $W^0_t$ as in Figure \ref{systemicRiskCommonNoise:SamplePaths}. We can observe a upwards shift in the measure flow, when compared with the case where the interaction is through the mean.
Intuitively, when every agent strives to be better than average, the population as a whole drifts upwards.

\subsection{Economic growth model}
\subsubsection{Problem formulation}
Our next example examines the limit of a stochastic economy with $N$ agents who make consumption-savings decisions over a finite time horizon, such as those described in \cite{achdou2022income}, when $N$ goes to infinity. Each agent starts with random initial capital $k^i$ and chooses their consumption rate to maximize a given utility. The capital dynamics depend on consumption, depreciation, and an endogenous interest rate determined by aggregate capital in the economy. This creates strategic interaction through the interest rate --- each individual agent's optimal consumption depends on the aggregate capital level, while the aggregate capital evolves based on all agents' consumption choices. We then consider the mean field limit with a continuum of agents distributed according to a probability measure flow. The Nash equilibrium is then characterized by an MV-FBSDE system. The model demonstrates how individual optimization and aggregate dynamics couple through the interest rate mechanism.

Consider an idealized economy where the $i$th agent capital dynamics, denoted $K_t^i$, is governed by two factors in addition to their consumption $c^i$:
a fixed depreciation rate $\delta$ and an endogenous interest rate $r_t$.
Moreover, it is subjected to an idiosyncratic noise $W^i$ and a common noise $W^0$ that affects all agents. The dynamics is described by the SDE:
\begin{align}
    \dK_t^i &=  \left( \left(r_t - \delta \right) K_t^i - c_t^i \right)\, \dt + \sigma( \rho\, \dW_t^0 + \sqrt{1-\rho^2}\, \dW^i_t), 
    &
    K_0^i = k^i.
\end{align}
We further assume that the economy's aggregate production $P_t$  is given by a function $F$ of the aggregate capital of the economy $\overline K_t$, that is,
\begin{equation}
    P_t = F(\overline K_t),\quad \text{where } \overline K_t = \frac{1}{N} \sum_{i = 1}^N K_t^i.
\end{equation}
In economic equilibrium, the interest rate $r_t$ is  given by the marginal effect of capital $\partial_K P$ in the aggregate production,
which itself is a function of the aggregate capital in the economy $\overline K_t$.
This is the source of the mean field interactions in this model.

The agents' consumption preferences are governed by the utility function $u(\cdot)$, 
and their preference for capital at the end of the time horizon are governed by a terminal utility $\psi(K_T^i)$.
Therefore, agents choose their consumption $c : [0,T] \times \Omega \longrightarrow C$, with $C \subset \R$, to maximize the following functional:
\begin{equation}
    J^i(k^i, c^i) = \E\left[\int_0^T u(c_t^i) \dt + \psi(K_T^i)\right],
\end{equation}
where $k^i$ is the initial capital and $K_T^i$ is their capital at time $T$ when following the control process $c^i$, and assumed to be sufficiently regular so the functional above is well defined. In all, the optimization problem faced by agent $i$ is
\begin{equation}\label{economic_example:N_player_game}
    \begin{cases}
        \displaystyle\max_{c^i \geq 0} \E \left[\int_0^T u(c^i_s) ds + \psi(K^i_T)\right],
        \\[0.25em]
        \text{subject to}
        \\[0.25em]
        d K_t^i = \left[\left( \partial_K F(\bar K_t) - \delta \right) K_t^i - c_t^i\right]\, dt + \sigma( \rho \,\dW_t^0 + \sqrt{1-\rho^2} \, \dW^i_t),
        & K_0^i=k^i.
    \end{cases}
\end{equation}

We next choose logarithmic utility $u(c) = \log(c)$, and then we assume the consumption takes values in the positive reals $C = \R_+$. Moreover, we consder quadratic terminal cost $\psi(K) = - \frac{1}{2} K^2$, and quadratic aggregation function $F(K) = \frac{C}{2} K^2$. With these choices, the optimal consumption is given by $c_t^\star = 1/Y_t$.
In the mean-field limit, we denote  the distribution of $K_t$ by $\mu_t$ and  the average capital by $S_t = \int k \mu_t({\rm d}k)$.
In this setting, the interest rate is given by $r_t = C \,S_t $.
Given an initial capital distribution $\mu_0$, the mean-field game is the solution to the MV-FBSDE system:
\begin{equation}\label{economic_example:ode_formulation}
    \begin{cases}
         \dK_t = \left(\left( r_t - \delta \right) K_t - \frac{1}{Y_t} \right)\, \dt + \sigma( \rho \,\dW_t^0 + \sqrt{1-\rho^2} \, \dW_t),
         &
         K_0 \sim \mu_0,
         \\[0.25em]
         \dY_t = - \left( \left(r_t - \delta \right) Y_t \right) \, \dt + Z_t\, \dW_t + Z^0_t \dW_t^0, 
         &Y_T =  - K_T.         
    \end{cases}
\end{equation}
The system implicitly describes the probability measure flow $\mu_t = \mathcal{L}(K_t | \mathcal{F}^0_t)$,
and the dynamics of this forward-backward system depend on $\mu_t$ through $r_t = C \, S_t = C \, \E[K_t | \mathcal{F}^0_t]$.
Note that the optimal consumption $c^*_t$ can be written as a function of $(t,K_t, r_t)$. Moreover, from the optimal consumption we can derive the \textit{marginal propensity to consume} (MPC), given by 
$$\partial_K c^*_t = -\frac{1}{Y_t^2} \frac{Z_t}{\sigma}.$$ 
This function describes, as a function of time, wealth and interest rate, how a increase in wealth is allocated between savings and consumption.

\subsubsection{Numerical simulation and parameters}

The initial condition follows a normal distribution with mean $0.5$ and standard deviation $0.5$. The system parameters are $C=1.5$, $\delta=0.1$, and $\sigma=0.1$.
We plot solutions with and without common noise in Figures \ref{economicGrowthModel:solutionCommonNoise} and \ref{economicGrowthModel:solutionNoCommonNoise}, respectively.

\begin{figure}[H]
\centering
\includegraphics[width=0.65\textwidth]{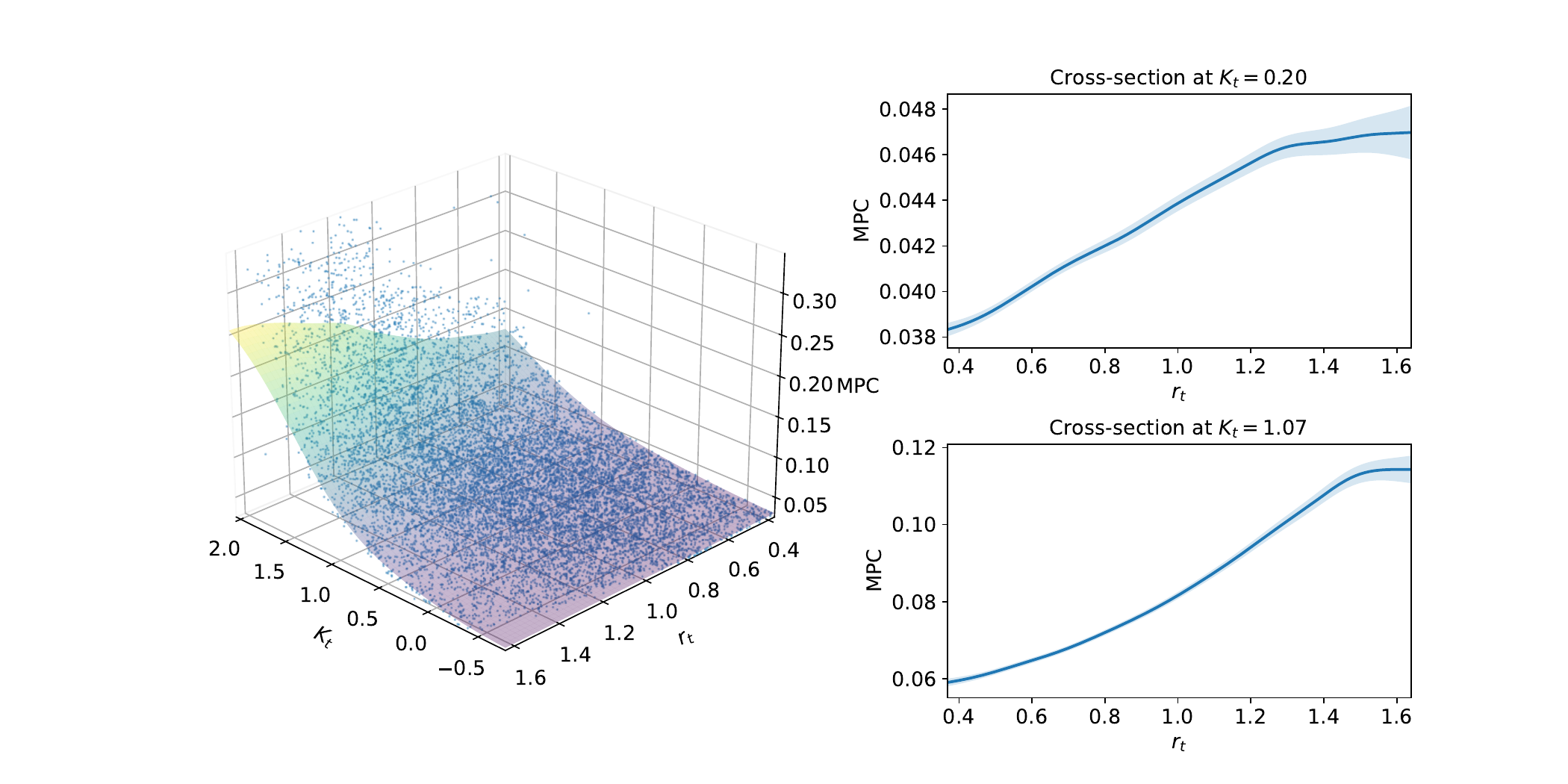}
\caption[Marginal Propensity to Consume at time 0.5]{Marginal propensity to consume as a function of $K_t$ and $r_t$, for time $t = 0.5$.}
\label{economicGrowthModel:mpc_05}
\includegraphics[width=0.65\textwidth]{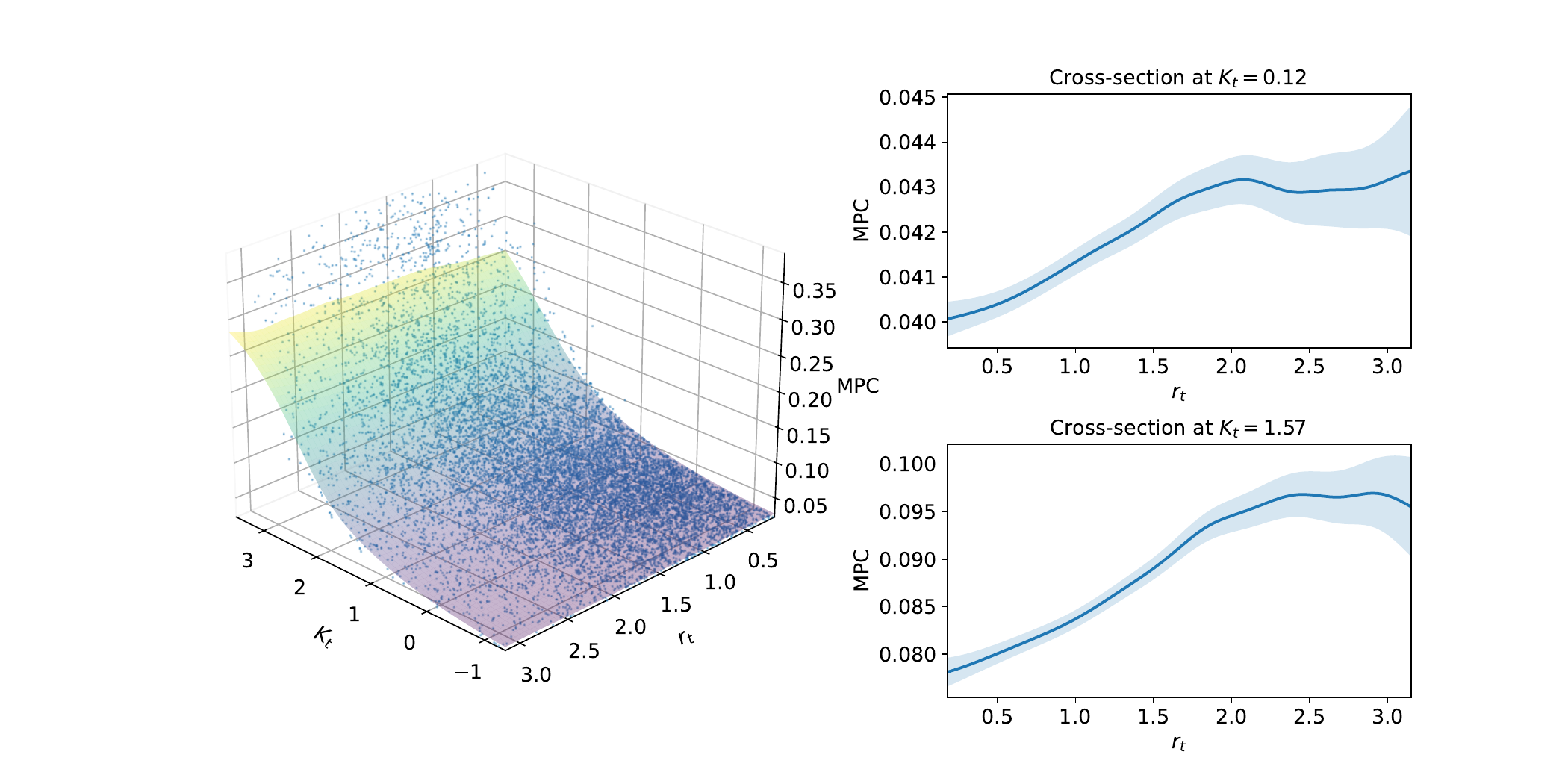}
\caption[Marginal Propensity to Consume at time 0.9]{Marginal propensity to consume as a function of $K_t$ and $r_t$, for time $t = 0.9$.}
\label{economicGrowthModel:mpc_09}
\end{figure}

In Figures \ref{economicGrowthModel:mpc_05} and \ref{economicGrowthModel:mpc_09}, we plot the marginal propensity to consume surface when subjected to common noise, as a function of the interest rate $r_t$ and capital $K_t$, at time $0.5$ and $0.9$ respectively. Note that at time $0.5$, MPC is increasing in $K_t$, but concave in $r_t$ - the intuition being that higher interest rates increase the payoff of saving behavior. However, the effect of the interest rate is reduced as we approach the end of the time interval.

\begin{minipage}[t]{0.5\linewidth}
\includegraphics[width=\textwidth]{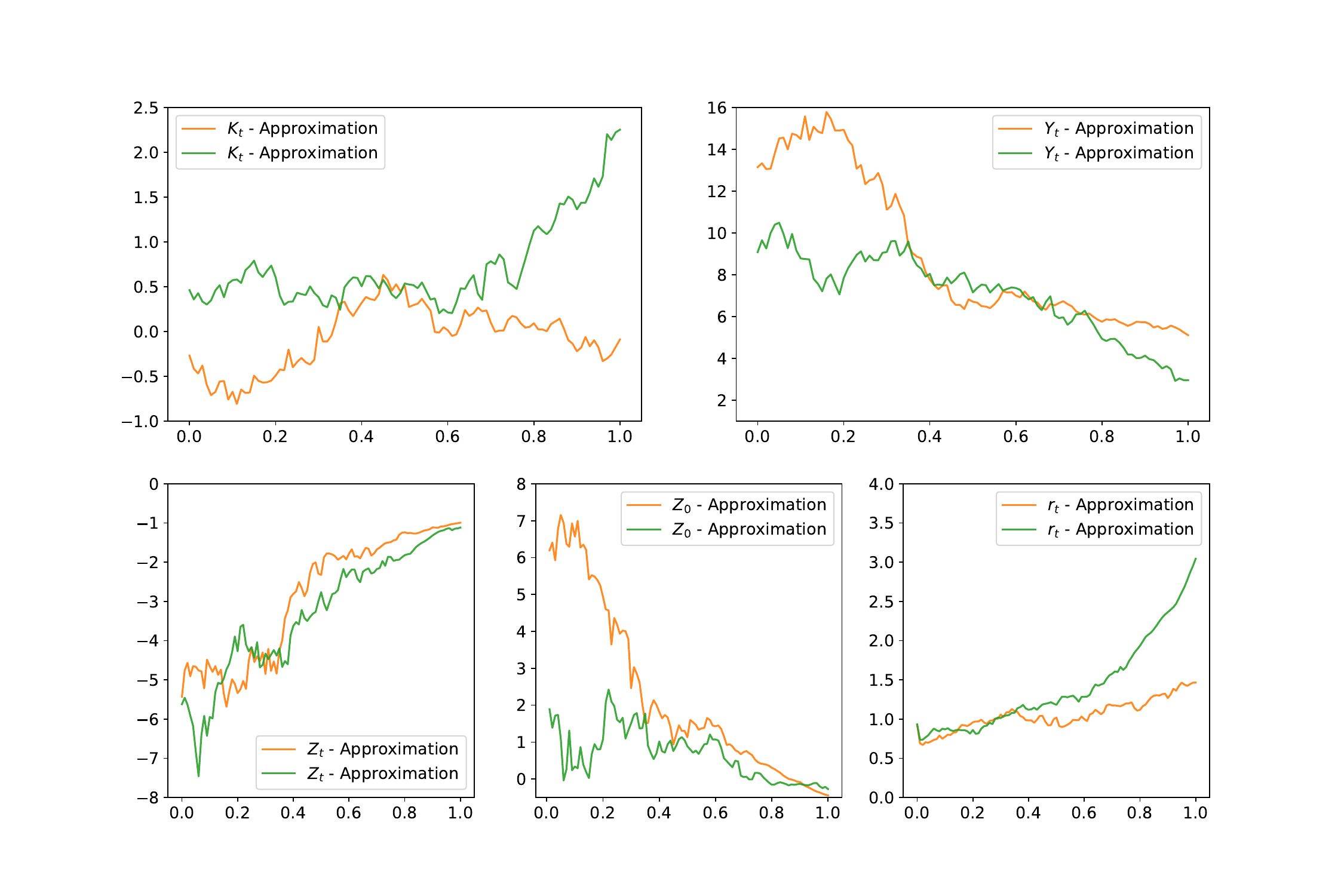}
\captionof{figure}{
Numerical solution of economic growth model with common noise.
}
\label{economicGrowthModel:solutionCommonNoise}
\end{minipage}
\begin{minipage}[t]{0.5\linewidth}
\includegraphics[width=\textwidth]{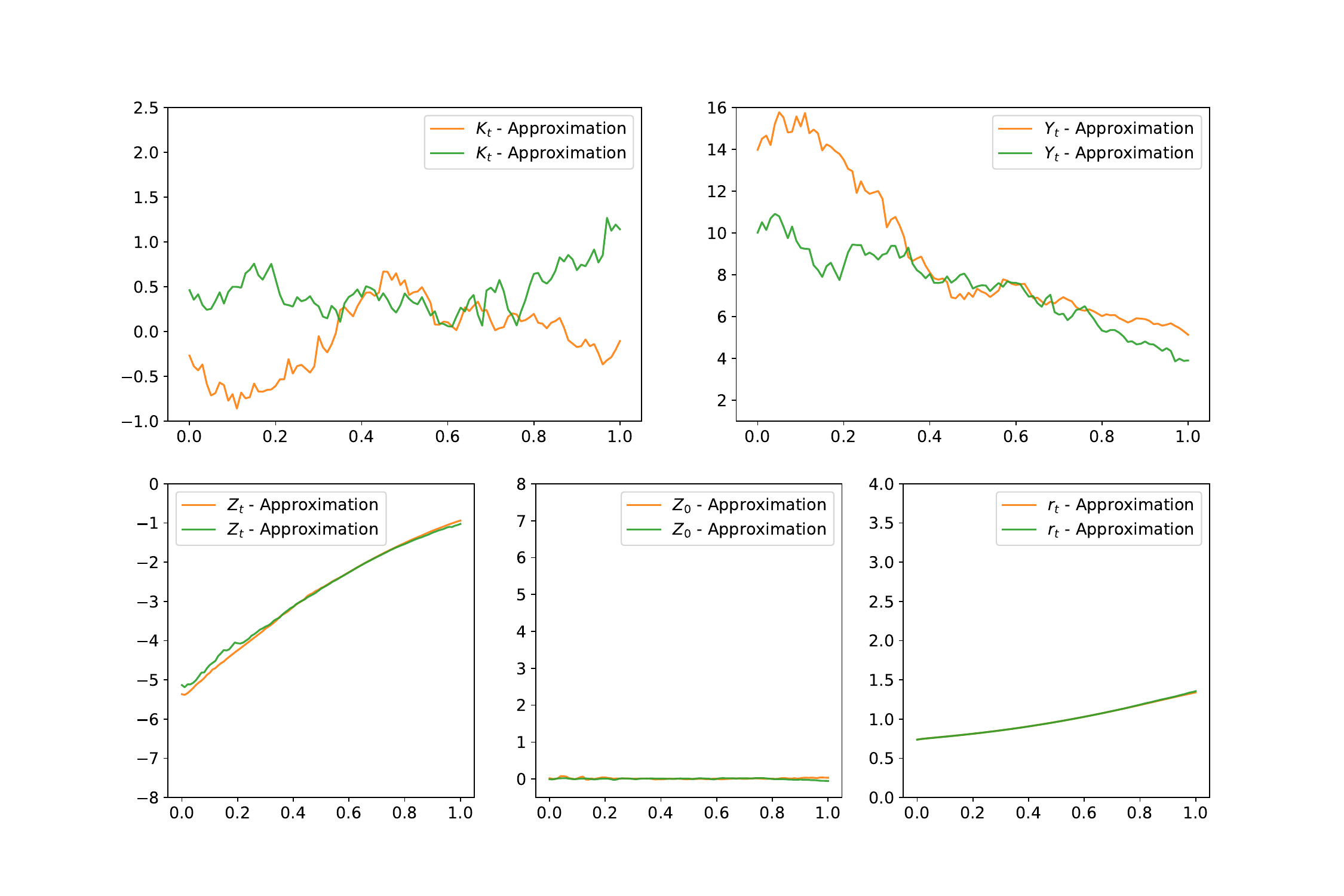}
\captionof{figure}{
Numerical solution of economic growth model without common noise. Note that the model learns that $Z_0$ is irrelevant in this setting.
}
\label{economicGrowthModel:solutionNoCommonNoise}
\end{minipage}

\section{Conclusion}\label{sec:conclusion}

We have presented a novel numerical method for solving McKean-Vlasov forward-backward stochastic differential equations with common noise. Our approach combines Picard iterations with elicitability principles and deep learning to address the two fundamental challenges in these systems: the forward-backward coupling and the dependence on the conditional measure flow. The key innovation of our method lies in the use of elicitability to compute conditional expectations, which enables us to handle the common noise case without resorting to computationally expensive nested Monte Carlo simulations.  

Our numerical experiments demonstrate the effectiveness of the proposed method across problems of varying complexity. The systemic risk banking model provides validation against known analytical solutions, confirming that our algorithm accurately recovers the true solution processes. The extension to quantile-mediated interactions showcases the flexibility of the elicitability framework in handling statistics beyond conditional means. Finally, the economic growth model illustrates the method's applicability to more complex, economically meaningful problems where analytical solutions are unavailable.

\section*{Acknowledgments}

SJ would like to acknowledge support from the Natural Sciences and Engineering Research Council of Canada through grant RGPIN-2024-04317. YS was supported by FAPERJ (Brasil) through the Jovem Cientista do Nosso Estado Program (E-26/201.375/2022 (272760)) and by CNPq (Brasil) through the Productivity in Research Scholarship (306695/2021-9). FA was supported by CAPES (Brazil) through Programa Institucional de Internacionalização (88887.939145/2024-00) and Programa Suporte à Pós-Graduação (88887.705168/2022-00), by FGV's School of Applied Mathematics and by UofT through a Research Assistant Award.

\printbibliography

\end{document}